# MAPP: a Scalable Multi-Agent Path Planning Algorithm with Tractability and Completeness Guarantees

**Ko-Hsin Cindy Wang**                                    CINDY.WANG@RSISE.ANU.EDU.AU
**Adi Botea**                                              ADI.BOTEA@NICTA.COM.AU
*NICTA & The Australian National University,*
*Canberra, Australia*

## Abstract

Multi-agent path planning is a challenging problem with numerous real-life applications. Running a *centralized* search such as A* in the combined state space of all units is complete and cost-optimal, but scales poorly, as the state space size is exponential in the number of mobile units. Traditional *decentralized* approaches, such as FAR and WHCA*, are faster and more scalable, being based on problem decomposition. However, such methods are incomplete and provide no guarantees with respect to the running time or the solution quality. They are not necessarily able to tell in a reasonable time whether they would succeed in finding a solution to a given instance.

We introduce MAPP, a tractable algorithm for multi-agent path planning on undirected graphs. We present a basic version and several extensions. They have low-polynomial worst-case upper bounds for the running time, the memory requirements, and the length of solutions. Even though all algorithmic versions are incomplete in the general case, each provides formal guarantees on problems it can solve. For each version, we discuss the algorithm's completeness with respect to clearly defined subclasses of instances.

Experiments were run on realistic game grid maps. MAPP solved 99.86% of all mobile units, which is 18–22% better than the percentage of FAR and WHCA*. MAPP marked 98.82% of all units as provably solvable during the first stage of plan computation. Parts of MAPP's computation can be re-used across instances on the same map. Speed-wise, MAPP is competitive or significantly faster than WHCA*, depending on whether MAPP performs all computations from scratch. When data that MAPP can re-use are preprocessed offline and readily available, MAPP is slower than the very fast FAR algorithm by a factor of 2.18 on average. MAPP's solutions are on average 20% longer than FAR's solutions and 7–31% longer than WHCA*'s solutions.

## 1. Introduction

Path planning is important in many real-life problems, including robotics, military operations, disaster rescue, logistics, and commercial games. Single-agent path planning, where the size of the state space is bounded by the size of the map, can be tackled with a search algorithm such as A* (Hart, Nilsson, & Raphael, 1968). However, when there are many units moving *simultaneously* inside a shared space, the problem becomes much harder. A *centralized* search from an initial state to a goal state is a difficult problem even inside a fully known, two-dimensional environment represented as a weighted graph, where one node can be occupied by exactly one unit at a time. Assuming that units have the same size, and each unit moves synchronously to an adjacent unoccupied node in one time step, the problem's state space grows exponentially in the number of mobile units. Existing hardness results have shown that it is NP-complete to decide if a solution of at most $k$ moves exists (Ratner & Warmuth, 1986), or to optimize the solution makespan (Surynek, 2010b). A version of the problem with one robot only and movable obstacles at several nodes, where either the robot or an obstacle can move to an adjacent vacant node per step, is also NP-complete (Papadimitriou, Raghavan, Sudan, & Tamaki, 1994). Yet another version of the problem, determining if a solution exists for moving two-dimensional rectangles of different sizes inside a box, has been shown to be PSPACE-hard, even without requiring optimality (Hopcroft, Schwartz, & Sharir, 1984). Despite its completeness and solution optimality guarantees, a centralized A* search has little practical value in a multi-agent path planning problem, being intractable even for relatively small maps and collections of mobile units.

Scalability to larger problems can be achieved with *decentralized* approaches, which decompose the global search into a series of smaller searches to significantly reduce computation. However, existing decentralized methods such as FAR (Wang & Botea, 2008) and WHCA* (Silver, 2006) are incomplete, and provide no formal criteria to distinguish between problem instances that can be successfully solved and other instances. Further, no guarantees are given with respect to the running time and the quality of the computed solutions.

In this work we present an algorithm that combines the strengths of both worlds: working well in practice *and* featuring theoretical tractability and partial completeness guarantees. We introduce MAPP, a tractable *multi-agent path planning* algorithm for undirected graphs. For each problem instance, MAPP systematically identifies a set of units, which can contain all units in the instance, that are guaranteed to be solved within low-polynomial time. For the sake of clarity we will distinguish between a basic version and a few extended versions of MAPP. MAPP provides formal guarantees for problems it can solve. The Basic MAPP algorithm is complete for a class of problems, called SLIDABLE, which we define in Section 3. Extended versions of the algorithm enlarge the completeness range, as discussed in Section 7, and improve solution length, as discussed in Section 8. We will also evaluate a version that attempts to solve all units, not only the provably solvable ones.

Given a problem with $m$ graph nodes and $n$ mobile units, MAPP's worst case performance for the running time is $O(m^2n^2)$, or even smaller (e.g., $O(\max(mn^2, m^2 \log m))$), depending on the assumptions on the input instance. The worst-case memory requirements are within $O(m^2n)$ or even $O(mn)$. An upper bound of the solution length, measured as the total number of moves, is in the order of $O(m^2n^2)$ or even $O(mn^2)$. See Section 6 for a detailed discussion.

MAPP keeps its running costs low by eliminating the need for replanning. A path $\pi(u)$ for each unit $u$ is computed at the beginning. No replanning is required at runtime. A *blank travel* idea, inspired from the way the blank moves around in sliding tile puzzles, is at the center of the algorithm. A unit $u$ can progress from its current location $l_i^u$ to the next location $l_{i+1}^u$ on its path $\pi(u)$ only if a *blank* is located there (i.e., $l_{i+1}^u$ is empty). Intuitively, if the next location is currently occupied by another unit, MAPP tries to bring a blank along an *alternate path*, outlined in bold in Figure 1, which connects $l_{i-1}^u$ and $l_{i+1}^u$ without passing through $l_i^u$. When possible, the blank is brought to $l_{i+1}^u$ by shifting units along the alternate path, just as the blank travels in a sliding tile puzzle. The ability to bring a blank to the next location is key to guarantee a unit's progress. Formal details are provided in Section 5.

We performed detailed experiments, evaluating different versions of MAPP and comparing MAPP with fast but incomplete methods such as FAR and WHCA* on grid maps. The results are presented in Section 9. The benchmark data (Wang & Botea, 2008) consist of 100 to 2000 mobile units uniformly randomly generated on 10 game maps, with 10 scenario instances per number of units of each map. We conclude that the extended MAPP has significantly better success ratio and scalability than state-of-the-art incomplete decentralized algorithms. In particular, MAPP solves a higher percentage of units even on crowded instances. Despite MAPP's incompleteness in the general case, the algorithm marks 98.82% of all units as *provably* solvable during the first stage of plan computation. When attempting to solve all units, not only the provably solvable ones, MAPP succeeds for 99.86% of all units. In comparison, FAR solved 81.87% of all units. WHCA* solved 77.84% (with diagonal moves allowed) and 80.87% (with no diagonal moves) of all units. Even in challenging instances with 2,000 mobile units on the maps, 92% to 99.7% mobile units in our test data fall within MAPP's completeness range (i.e., they are provably solvable). In terms of the percentage of fully solved instances, a version of MAPP that attempts to solve all units, not only those that are provably solvable, is successful in 84.5% of all instances. This is significantly better than FAR (70.6%), WHCA* with no diagonal moves (58.3%), and WHCA* with diagonals (71.25%).

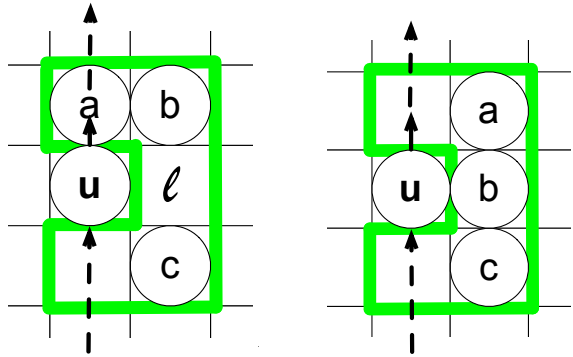

Figure 1: At the left, unit $u$ is blocked by $a$. A blank is found at location $l$ along the alternate path, which is marked with a bold contour. At the right: by sliding $b$ and then $a$ along the alternate path, the blank is brought to $l_{i+1}^u$. For the sake of clarity and simplicity, we illustrate examples in a four-connected grid world.

Parts of MAPP's computation can be re-used across instances on the same map. On instances solved by all algorithms, MAPP is competitive in speed or significantly faster than WHCA*, depending on whether MAPP performs all computations from scratch. When such re-usable data are available, MAPP is slower than the very fast FAR algorithm by a factor of 2.18 on average. MAPP's solutions reported here are on average 20% longer than FAR's solutions and 7–31% longer than WHCA*'s solutions.

Parts of this work have been reported in shorter conference papers as follows. A theoretical description of Basic MAPP, with no experiments, is provided in an earlier paper (Wang & Botea, 2009). A brief overview of MAPP extensions and a brief summary of initial results are the topic of a two-page paper (Wang & Botea, 2010). New material added in the current paper includes a detailed algorithmic description of the enhancements to Basic MAPP and formal proofs for the algorithms' properties. We also provide a comprehensive empirical analysis of enhanced MAPP, with several additional experiments.

The rest of this paper is structured as follows. Next we briefly overview related work. Then, we state our problem definition in Section 3. Sections 4–6 focus on Basic MAPP. Sections 7 and 8 cover enhancements to the Basic MAPP algorithm, extending its completeness range (Section 7), and improving the quality of plans and also the running time (Section 8). An empirical evaluation is the topic of Section 9. The last part contains conclusions and future work ideas.

## 2. Related Work

Finding a shortest path that connects a single pair of start-target points on a known, finite map can be optimally solved with the A* algorithm (Hart et al., 1968). The extension to path planning for multiple simultaneously moving units, with distinct start and target positions, introduces potential collisions due to the physical constraint that one location can only be occupied by one unit at a time. Units have to interact and share information with other units in their path planning, making the problem more complex.

In multi-agent path planning, a centralized A* performs a single global search in the combined state space $L_1 \times L_2 \times \cdots \times L_n$ for $n$ units, where $L_i$ is the set of possible locations of unit $i$. Centralized A* plans the paths for all units simultaneously, finding a joint plan containing all units' actions (waits as well as moves). It retains the optimality and completeness guarantees of A*, but has a prohibitively large state space of $O(m^n)$ states, for $n$ units on a graph with $m$ nodes. Moreover, most of the search nodes generated are 'unpromising', taking some units farther from goal (Standley, 2010). This poses a strong limiting factor on problems that a centralized A* can solve in practice. On the other hand, a purely decentralized method, Local Repair A* (Lra*) (Stout, 1996) first plans each unit's path independently with A*. Then, during execution, Lra* replans by additional independent A* searches every time a collision occurs. In a good case, Lra* can significantly reduce computations to $O(mn)$. However, it can also generate cycles between units, and is unable to prevent bottlenecks. These problems have been discussed by Silver (2005), Bulitko, Sturtevant, Lu, and Yau (2007), Pottinger (1999), and Zelinsky (1992). In such cases, Lra* exhibits a significant increase in running time and may not terminate. Therefore, all of the straightforward extensions of single-agent A* outlined above have strong limitations in practice.

Traditionally, multi-agent path planning took a centralised or a decentralised approach (Latombe, 1991; Choset et al., 2005). A *centralized* approach plans globally, sharing information centrally, such as using a potential field (Barraquand, Langlois, & Latombe, 1991). By contrast, a *decentralized* approach decomposes the problem into a series of smaller subproblems, typically first computing the units' paths individually, ignoring all other units, then handling the interactions online. Examples in robotics include computing velocity profiles to avoid collisions with other units (Kant & Zucker, 1986), or pre-assigning priorities to process robots one by one (Erdmann & Lozano-Perez, 1986). Recent algorithms can also use a combination of the two approaches. For instance, the Biased Cost Pathfinding (BCP) technique (Geramifard, Chubak, & Bulitko, 2006) generalised the notion of centralized planning to a central decision maker that resolves collision points on paths that were pre-computed independently per unit, by replanning colliding units around a highest-priority unit. To avoid excessively long (or even potentially unbounded) conflict resolutions, a limit on planning time is set. BCP returns paths with the fewest collisions within that time. The algorithm was shown to work well in small-scale gridworld scenarios, but it is not complete or optimal in the general case. Standley's (2010) algorithm, on the other hand, improved the "standard" centralized search whilst preserving both optimality and completeness. His new state space representation incorporates the next move assignments of every unit into each state, and decomposes a timestep from advancing all units to advancing units one by one in a fixed ordering. Thus the branching factor is reduced from $9^n$ to 9, while increasing the depth of the search by a factor of $n$. This technique gen-erates no more than $9nt$ state nodes with a perfect heuristic ($t$ being the number of timesteps in the optimal solution). In practice, this operator decomposition technique (OD) is still often intractable, producing a lower exponential search space than the standard joint search space. Recognising that it is much cheaper to perform several independent searches than one global search, Standley also decoupled planning for non-interfering subgroups of units after an independence detection (ID). Each group is then solved centrally such that optimality of the overall solution is still guaranteed. The fully developed hybrid algorithm, OD+ID, uses operator decomposition to improve the centralized planning of non-independent subproblems. Nonetheless, the optimality requirement is costly in practice. Planning time is still dominated by the largest subgroup of units. As the number of units increases, they are less likely to be independent as their paths unavoidably overlap, so the subgroups are expected to increase in size too. Standley's (2010) experiments showed that the incomplete algorithm HCA* (Silver, 2005) actually solved more instances. Furthermore, these are relatively small problems compared to our experiments (Wang & Botea, 2008, 2010), having at least 2 orders of magnitude fewer agents (between 2–60 units), and much smaller maps, with 1 to 2 orders of magnitude fewer tiles (approximately 819 tiles).

Therefore, methods for tackling larger problems take a decentralized approach, and are usually suboptimal in nature. In general, giving up optimality reduces computation significantly. Decentralized path planning is often much faster, and scales up to much larger problems, but yields suboptimal solutions and provides no completeness guarantees. Recent work on grid maps include WHCA* (Silver, 2006), which uses a 3-dimensional temporal-spatial reservation table and performs a series of windowed forward searches on each unit, based on a true distance heuristic obtained from an initial backward A* search from each target. In the FAR algorithm (Wang & Botea, 2008), units follow a flow annotation on the map when planning and moving, repairing plans locally using heuristic procedures to break deadlocks. Other 'flow' related ideas include Jansen and Sturtevant's (2008) direction map for sharing information about units' directions of travel, so later units can follow the movement of earlier ones, with the improved coherence leading to reduced collisions. Methods such as these scale up to instances with the number of units well beyond the capabilities of centralized search. However, as mentioned earlier, these methods have no known formal characterizations of their running time, memory requirements, and the quality of their solutions in the worst case. They lack the ability to answer in a reasonable bounded time whether a given problem would be successfully solved, which is always important in the case of incomplete algorithms.

In practice, both traditional approaches to multi-agent pathfinding have serious drawbacks, with the inherent trade-off between scalability, optimality and completeness. Recently, a body of work has begun to bridge the gap between the two, by addressing both completeness and tractability issues hand in hand, in a *bounded suboptimal* approach. Ryan (2008) introduced a complete method that combines multi-agent path planning with hierarchical planning on search graphs with specific substructures such as stacks, halls, cliques and rings. For example, a stack is a narrow corridor with only one entrance, which is placed at one end of the stack. Many maps, including the game maps used in our experiments, seem not to allow an efficient decomposition into stacks, halls, cliques and rings. BIBOX (Surynek, 2009b) solves problems with at least 2 unoccupied vertices on a biconnected graph. In the worst case, the number of steps is cubic in the number of nodes. BIBOX was later extended to work with just 1 unoccupied vertex necessary (Surynek, 2009a). Because of the densely populated problems that the algorithm was designed for, Surynek (2010a) has expressed that BIBOX does not target computer game scenarios, where there are normally a lot fewer units than locations on the map. BIBOX is suited for multi-robot scenarios such as automatic packages inside a warehouse (Surynek, 2010c). Bibox-$\theta$ (Surynek, 2009a), that requires only 1 unoccupied node, was shown to run significantly faster and have significantly shorter solutions than Kornhauser, Miller, and Spirakis's (1984) algorithm for their related 'pebble coordination game'. We performed a quick evaluation of BIBOX using the code obtained from the author. We found that, on graphs at an order of magnitude smaller than our game maps, BIBOX exhibits a fast-growing runtime (e.g., more than 10 minutes for a graph with 2500 locations) and long solutions, with millions of moves. Part of the explanation is that BIBOX builds its instances to be very crowded. In our understanding, BIBOX was designed to solve very crowded instances, not necessarily to efficiently solve instances with significantly fewer units than locations.

## 3. Problem Statement

An *instance* is characterized by a graph representation of a map, and a non-empty collection of mobile units $U$. Units are homogeneous in speed and size. Each unit $u \in U$ has an associated start-target pair $(s_u, t_u)$. All units have distinct starting and target positions. The objective is to navigate all units from their start positions to the targets while avoiding all fixed and mobile obstacles. A *state* contains the positions of all units at a given time. Our work assumes undirected weighted graphs where each unit occupies exactly one node at a time, and can move to an unoccupied neighbour node. The time is discretized and one or more units can move synchronously at each time step. Travelling along an edge does not depend on or interfere with the rest of the problem, except for the two nodes connected by that edge.

Several methods exist to abstract a problem map into a search graph, including navigation meshes (Tozour, 2002), visibility points (Rabin, 2000), and quadtrees (Samet, 1988). However, a graph abstraction that generates too few nodes, such as a visibility graph, may render a multi-agent pathfinding problem unsolvable, even though it works for the single agent case. On the other hand, a search graph obtained from imposing a regular grid contains more nodes, covering all locations of the traversable space, and offers more path options to avoid collisions between units. Hence, grid maps, besides being very popular and easy to implement, are more suitable to multi-agent problems. For clarity and practicality, we focus on grid maps in our examples and experiments. Nonetheless, the conditions and algorithmic definitions for MAPP, which we introduce in the next few sections, are not specific to regular grid maps. In our illustrated examples, we assume that only straight moves in the four cardinal directions can be performed (4 connected grid). Restricting the movements from 8 directions (cardinal + diagonals) to 4 cardinal directions has no negative impact on completeness. Since the standard practice is to allow a diagonal move only if an equivalent (but longer) two-move path exists, for every solution that allows diagonal moves, there is a solution with only cardinal moves. Therefore, any problem with diagonal moves can be reduced to a problem with only straight moves, at the price of possibly taking longer paths. Introducing diagonal moves could reduce the path length, but has the potential drawback of blocking units more often than straight moves on crowded maps. Whether there is enough clearance to make a diagonal move depends on the other two adjacent nodes (i.e., the other two tiles sharing the common corner on the grid), since it is physically impossible to squeeze through two units.

## 4. The SLIDABLE Class of Instances

We introduce a subclass of instances for which Basic MAPP will be shown to be complete.

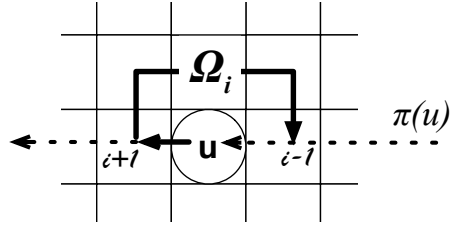

Figure 2: An example of an alternate path, $\Omega_i$, connecting locations $l_{i-1}^u$ and $l_{i+1}^u$ (denoted as $i-1$ and $i+1$ in the picture) that belong to the precomputed path $\pi(u)$ of unit $u$.

**Definition 1** (SLIDABLE unit and SLIDABLE instance). *A mobile unit $u$ is SLIDABLE iff a path $\pi(u) = (l_0^u, l_1^u, \ldots, l_{|\pi(u)|}^u)$ of nodes exists, where $l_0^u = s_u$, $l_{|\pi(u)|}^u = t_u$, such that all the following conditions are met:*

1. Alternate connectivity. *For each three consecutive locations $l_{i-1}^u, l_i^u, l_{i+1}^u$ on $\pi(u)$, except the last triple ending with $t_u$, i.e. for $0 < i < |\pi(u)| - 1$, an alternate path $\Omega_i^u$ exists between $l_{i-1}^u$ and $l_{i+1}^u$ that does not go through $l_i^u$. See Figure 2 for an example.*

2. Initial blank. *In the initial state, $l_1^u$ is blank (i.e. unoccupied).*

3. Target isolation. *No target interferes with the $\pi$ or $\Omega$-paths of the other units. More formally, both of the following hold for $t_u$:*

    (a) *$(\forall v \in U \setminus \{u\}) : t_u \notin \pi(v)$; and*

    (b) *$(\forall v \in U, \forall i \in \{1, \ldots, |\pi(v)| - 1\}) : t_u \notin \Omega_i^v$.*

*An instance belongs to the class SLIDABLE iff all units $u \in U$ are SLIDABLE.*

The three conditions can be verified in polynomial time. The verification includes attempting to compute the $\pi$ and $\Omega$ paths for each unit. Since each state space that A* has to explore here is linear in $m$, each A* search time is polynomial in $m$. The checks for a blank location in the first step, and for not passing through other targets, are trivial. The process that checks the SLIDABLE conditions serves for an important additional purpose. By the time the checks succeed and an instance is known to belong to SLIDABLE, we have completed all the search that is needed to solve the instance. The remaining part of the algorithm will simply tell units when to wait, when to move forward, and when to move backwards along the already computed $\pi$ and $\Omega$ paths.

Notice that the three conditions are not restricted to grid maps only. They work on the standard assumption that one graph node can only be occupied by one unit at a time, and that moving along an edge neither depends nor interferes with other parts of the graph except for two nodes at the ends of that edge.

---

**Algorithm 1** Overview of MAPP.

---

1: **for** each $u \in U$ **do**
2:    compute $\pi(u)$ and $\Omega$'s (as needed) from $s_u$ to $t_u$
3:    **if** SLIDABLE conditions hold **then**
4:       mark $u$ as SLIDABLE
5: initialize $A$ as the set of SLIDABLE units {optional: make *all* units active, as discussed in text}
6: **while** $A \neq \emptyset$ **do**
7:    do progression step
8:    do repositioning step if needed

---

## 5. Basic MAPP

We present the basic version of the MAPP algorithm, which is complete on the SLIDABLE class of problems. A main feature of Basic MAPP (and its extensions presented in Sections 7 and 8) is that it is deadlock-free and cycle-free, due to a total ordering of active units. Units of lower priority do not interfere with the ability of higher priority units to advance.

As illustrated in Algorithm 1, for each problem instance, MAPP starts by computing a path $\pi(u)$ for each unit $u$ to its target (goal), constructing and caching alternate paths $\Omega$ along the way. Note that all paths $\pi$ and alternate paths $\Omega$ need to satisfy the conditions in Definition 1. If the *for* loop in lines 1–4 succeeds for all units, MAPP can tell that the instance at hand belongs to SLIDABLE, for which MAPP is complete.

If only a subset of units are marked as SLIDABLE, MAPP is guaranteed to solve them. This is equivalent to solving a smaller instance that is SLIDABLE. Optionally, MAPP can attempt to solve the remaining units as well, by adding them to the set of active units but giving them a lower priority than SLIDABLE units. It is important to stress out that, in the remaining part of the paper, the implicit assumption is that MAPP attempts to solve only the provably solvable units, unless we explicitly state the opposite. In the experiments section, however, we discuss both options.

The set of SLIDABLE units is partitioned into a subset $S$ of *solved units* that have already reached their targets, and a subset $A$ of *active units*. Initially, all units are active. In the SLIDABLE class, after becoming solved, units do not interfere with the rest of the problem (as ensured by the target isolation condition). As shown later, in Basic MAPP solved units never become active again, and do not have to be considered in the remaining part of the solving process.

**Definition 2.** *The* advancing condition *of an active unit $u$ is satisfied iff its current position, $pos(u)$, belongs to the path $\pi(u)$ and the next location on the path is blank.*

**Definition 3.** *A state is* well positioned *iff all active units have their advancing condition satisfied.*

Lines 6–8 in Algorithm 1 describe a series of two-step iterations. A *progression step* advances active units towards their targets. As shown later, each progression step brings at least one active unit to its target, shrinking the active set $A$ and ensuring that the algorithm terminates, reaching the state where all units are solved. A progression could result in breaking the advancing condition of one or more active units, if any remain. The objective of a *repositioning step* is to ensure that each active unit has its advancing condition satisfied before starting the next progression step. Note that a repositioning step is necessary after every progression step except for the last.

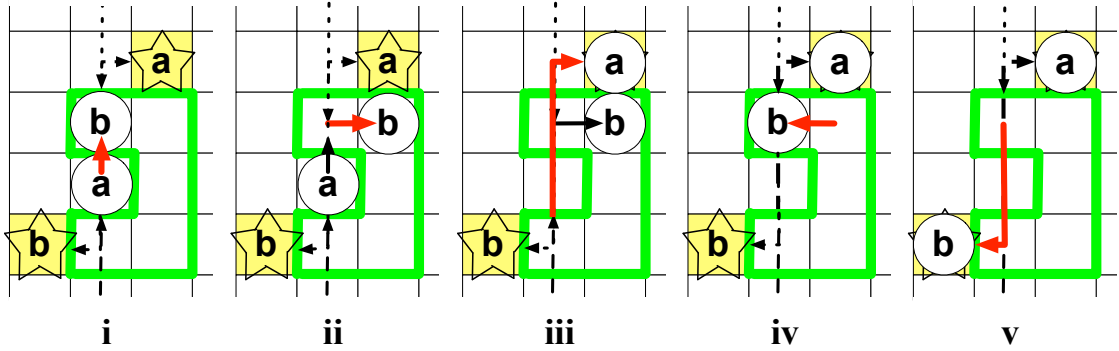

Figure 3: Example of how MAPP works.

## 5.1 Example

A simple example of how MAPP works is illustrated in Figure 3. There are two units, $a$ and $b$. MAPP uses a total ordering of the active units in each progression step (Section 5.3). Here, $a$ has a higher priority than $b$. The targets of $a$ and $b$ are drawn as stars. In Figure 3 (i), as $a$ and $b$ progress towards their targets, $a$ becomes blocked by $b$. In (ii), a blank is brought in front of $a$ by sliding $b$ down $\Omega_i^a$ (outlined in bold); as a side effect, $b$ is pushed off its path. At the end of the current progression step (iii), $a$ reaches its target. In the repositioning step (iv), since $a$ is already solved, $a$'s moves are ignored. Repositioning undoes $b$'s moves until $b$ is back on its path and it has a blank in front of it. Now $b$'s advancing condition is restored and therefore the global state in this example is well positioned. In the next progression step (v), $b$ reaches its target. The algorithm terminates.

## 5.2 Path Computation

For each problem instance, we compute each path $\pi(u)$ individually. The paths $\pi(u)$ are fixed throughout the solving process. To ensure that paths satisfy the alternate connectivity condition (Definition 1), we modify the standard A* algorithm as follows. When expanding a node $x'$, a neighbour $x''$ is added to the open list only if there is an alternate path between $x''$ and $x$, the parent of $x'$. By this process we compute each path $\pi(u)$ and its family of alternate paths $\Omega$ simultaneously. To give each neighbour $x''$ of the node $x'$ a chance to be added to the open list, node $x'$ might have to be expanded at most three times, once per possible parent $x$. Therefore, $O(m)$ node expansions are required by A* search to find each $\pi$ path, where $m$ is the number of locations on the map. Equivalently, computing a $\pi$ path could also be seen as a standard A* search in an extended space of pairs of neighbouring nodes (at most four nodes are created in the extended space for each original node).

Since alternate paths depend only on the triple locations, not the unit, we can re-use this information when planning paths for all units of the same problem. This means that the alternate path for any set of three adjacent tiles on the map is computed at most once per problem instance, and cached for later use. Given a location $l$ on a grid map, there are at most eight locations that could be on a path two moves away on a four-connect grid. As shown in Figure 4a, these eight locations form a diamond shape around $l$. For each of the four locations that are on a straight line from $l$

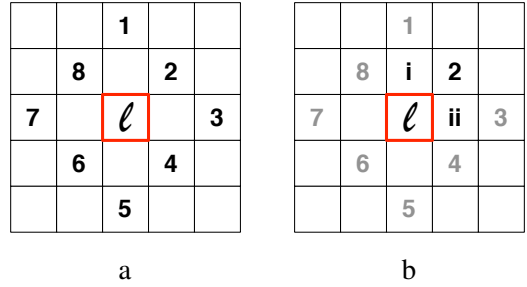

a         b

Figure 4: (a) The eight locations two moves away from $l$. (b) Two two-move paths from $l$ to location 2 go through $i$ and $ii$.

(locations 1, 3, 5, 7), we precompute an alternate path that avoids the in-between location and any targets. For each of the other four locations (labeled 2, 4, 6, 8), we need to compute (at most) two alternate paths. For example, there are two possible paths between $l$ and 2 that are two moves long: through $i$ or $ii$ (Figure 4b). We need one alternate path to avoid each intermediate location, $i$ and $ii$. In summary, we precompute at most 12 paths for each $l$. For at most $m$ locations on a map, we need at most $\frac{12m}{2} = 6m$ alternate paths (only one computation for each triple, since an alternate path connects its two endpoints both ways).

A possible optimization is to reuse alternate paths across SLIDABLE instances on the same map. Alternate paths that overlap targets in the new instance need to be re-computed. We discuss this option in the experiments section.

### 5.3 Progression

Algorithm 2 shows the progression step in pseudocode. At each iteration of the outer loop, active units attempt to progress by one move towards their targets. They are processed in order (line 2). If unit $v$ is processed before unit $w$, we say that $v$ has a higher priority and write $v < w$. The ordering is fixed inside a progression step, but it may change from one progression step to another. The actual ordering affects neither the correctness nor the completeness of the method, but it may impact the speed and the solution length. The ordering of units can be chosen heuristically, e.g. giving higher priority to units that are closer to target. Thus, these units could get to their target more quickly, and once solved they are out of the way of the remaining units in the problem.

To ensure that lower priority units do not harm the ability of higher priority units to progress, we introduce the notion of a private zone. We will see in Algorithm 2 that a unit cannot cause moves that will occupy the private zone of a higher-priority unit.[1] Given a unit $u$, let $pos(u)$ be its current position, and let $int(\pi(u)) = \{l_1^u, \dots, l_{|\pi(u)|-1}^u\}$ be the interior of its precomputed path $\pi(u)$. As shown in Algorithm 2, a unit $u$ might get pushed off its precomputed path, in which case $pos(u) \notin \pi(u)$.

**Definition 4.** *The* private zone, $\zeta(u)$, *of a unit* $u$ *is* $\zeta(u) = \{l_{i-1}^u, l_i^u\}$ *if* $pos(u) = l_i^u \in int(\pi(u))$. *Otherwise,* $\zeta(u) = \{pos(u)\}$. *In other words, the private zone includes the current location of the*

---

1. A move caused by a unit $u$ is either a move of $u$ along its own $\pi(u)$ path, or a move of a different unit $w$, which is being pushed around by $u$ as a side effect of blank travel.

---

**Algorithm 2** Progression step.

---

 1: **while** changes occur **do**
 2:     **for** each $u \in A$ in order **do**
 3:         **if** pos$(u) \notin \pi(u)$ **then**
 4:             do nothing {$u$ has been pushed off track as a result of blank travel}
 5:         **else if** $u$ has already visited $l_{i+1}^u$ in current progression step **then**
 6:             do nothing
 7:         **else if** the next location, $l_{i+1}^u$, belongs to the private zone of a higher priority unit, i.e.
                 $\exists v < u : l_{i+1}^u \in \zeta(v)$ **then**
 8:             do nothing {wait until $l_{i+1}^u$ is released by $v$}
 9:         **else if** $l_{i+1}^u$ is blank **then**
10:             move $u$ to $l_{i+1}^u$
11:         **else if** can bring blank to $l_{i+1}^u$ **then**
12:             bring blank to $l_{i+1}^u$
13:             move $u$ to $l_{i+1}^u$
14:         **else**
15:             do nothing

---

*unit. In addition, when the unit is on its pre-computed path but not on the start position, the location behind the unit belongs to the private zone as well.*

Lines 3–15 in Algorithm 2 show the processing of $u$, the active unit at hand. If $u$ has been pushed off its precomputed path, then no action is taken (lines 3–4). Lines 5 and 6 cover the situation when unit $u$ has been pushed around (via blank travel) by higher-priority units back to a location on $\pi(u)$ already visited in the current progression step. In such a case, $u$ doesn't attempt to travel again on a previously traversed portion of its path, ensuring that the bounds on the total travelled distance introduced later hold. If $u$ is on its path but the next location $l_{i+1}^u$ is currently blocked by a higher-priority unit $v$, then no action is taken (lines 7–8). Otherwise, if the next location $l_{i+1}^u$ is available, $u$ moves there (lines 9–10). Finally, if $l_{i+1}^u$ is occupied by a smaller-priority unit, an attempt is made to first bring a blank to $l_{i+1}^u$ and then have $u$ move there (lines 11–13). When $u$ moves to a new location $l_{i+1}^u$ (lines 10 and 13), a test is performed to check if $l_{i+1}^u$ is the target location of $u$. If this is the case, then $u$ is marked as solved by removing it from $A$ and adding it to $S$, the set of solved units.

Bringing a blank to $l_{i+1}^u$ (lines 11 and 12) was illustrated in Figure 1. Here we discuss the process in more detail. A location $l \in \Omega_i^u$ is sought with the following properties: (1) $l$ is blank, (2) none of the locations from $l$ to $l_{i+1}^u$ (inclusive) along $\Omega_i^u$ belongs to the private zone of a higher-priority unit, and (3) $l$ is the closest (along $\Omega_i^u$) to $l_{i+1}^u$ with this property. If such a location $l$ is found, then the test on line 11 succeeds. The actual travel of the blank from $l$ to $l_{i+1}^u$ along $\Omega_i^u$ (line 12) is identical to the movement of tiles in a sliding-tile puzzle. Figure 1 shows an example before and after blank traveling. The intuition behind seeking a blank along $\Omega_i^u$ is that, often, $l_{i-1}^u$ remains blank during the time interval after $u$ advances from $l_{i-1}^u$ to $l_i^u$ and until the test on line 11 is performed. This is guaranteed to always hold in the case of the active unit with the highest priority, which we call the master unit.

Let us introduce and characterize the behaviour of the master unit more formally. At the beginning of a progression step, one master unit $\bar{u}$ is selected. It is the unit with the highest priority among the units that are active at the beginning of the progression step. The status of being the master unit is preserved during the entire progression step, even after $\bar{u}$ becomes solved. At the beginning of the next progression step, a new master unit will be selected among the remaining active units.

**Lemma 5.** *The master unit $\bar{u}$ can always bring a blank to its front, if it needs one.*

*Proof.* Since $\bar{u}$'s previous location, $l_{i-1}^{\bar{u}}$, belongs to its private zone, $\zeta(\bar{u})$, and no other unit can move into the private zone of the highest priority unit, $\bar{u}$ is guaranteed to always find a blank at $l_{i-1}^{\bar{u}}$. Moreover, no location along $\Omega_i^{\bar{u}}$ from $l_{i-1}^{\bar{u}}$ to $l_{i+1}^{\bar{u}}$ can belong to the private zone of a higher priority unit since there are no units with a higher priority. Note also that $\Omega_i^{\bar{u}}$ is free of physical obstacles by construction. So it must be possible for the blank to travel from $l_{i-1}^{\bar{u}}$ to $l_{i+1}^{\bar{u}}$. $\qquad\square$

**Lemma 6.** *The master unit $\bar{u}$ is never pushed off its $\pi$-path.*

*Proof.* If $\bar{u}$ is pushed off $\pi(\bar{u})$ in blank travelling performed by another unit, it contradicts with $\bar{u}$ being the highest priority unit. $\qquad\square$

**Theorem 7.** *As long as the master unit $\bar{u}$ is not solved, it is guaranteed to advance along $\pi(\bar{u})$ at each iteration of the outer ("while") loop in Algorithm 2. By the end of the current progression step, at least $\bar{u}$ has reached its target.*

*Proof.* Using the previous two lemmas, it is easy to check that $\bar{u}$ never enters a "do nothing" line in Algorithm 2. Similar to Lemma 6, $\bar{u}$ is never pushed and cannot revisit a previous location. Also, since $\bar{u}$ has the highest priority, its next location cannot be held in the private zone of another unit. Hence, $\bar{u}$'s progress to its target is guaranteed. $\qquad\square$

The following result is useful to ensure that a progression step always terminates, either in a state where all units are solved or in a state where all remaining active units are stuck.

**Theorem 8.** *Algorithm 2 generates no cycles (i.e., no repetitions of the global state).*

*Proof.* We show a proof by contradiction. Assume that there are cycles. Consider a cycle and the active unit $u$ in the cycle that has the highest priority. Since no other unit in the cycle dominates $u$, it means that the movements of $u$ cannot be part of a blank travel triggered by a higher priority unit. Therefore, the movements of $u$ are a result of either line 10 or line 13. That is, all $u$'s moves are along its path $\pi(u)$. Since $\pi(u)$ contains no cycles, $u$ cannot run in a cycle. $\qquad\square$

## 5.4 Repositioning

By the end of a progression step, some of the remaining active units (if any are left) have their advancing condition broken. Recall that this happens for a unit $u$ when either $\mathrm{pos}(u) \notin \pi(u)$ or $u$ is placed on its precomputed path but the next location on the path is not blank. A repositioning step ensures that a well positioned state is reached (i.e., all active units have the advancing condition satisfied) before starting the next progression step.

A simple and computationally efficient method to perform repositioning is to undo a block of the most recent moves performed in the preceding progression step. Undoing a move means carrying out the reverse move. Solved units are not affected. For those remaining active units, we undo their moves, in reverse global order, until a well positioned state is encountered. We call this strategy *reverse repositioning*. An example is provided in Section 5.1.

**Proposition 9.** *If the reverse repositioning strategy is used at line 7 of Algorithm 1 (when needed), then all progression steps start from a well positioned state.*

*Proof.* This lemma can be proven by induction on the iteration number $j$ in Algorithm 1. Since the initial state is well positioned (this follows easily from Definitions 1 and 3), the proof for $j = 1$ is trivial. Assume that a repositioning step is performed before starting the iteration $j + 1$. In the worst case, reverse repositioning undoes all the moves of the remaining active units (but not the moves of the units that have become solved), back to their original positions at the beginning of $j$-th progression step. In other words, we reach a state $s$ that is similar to the state $s'$ at the beginning of the previous progression step, except that more units are on their targets in $s$. Since $s'$ is well positioned (according to the induction step), it follows easily that $s$ is well positioned too. □

## 6. Worst-case and Best-case Analysis

We give here bounds on the runtime, memory usage, and solution length for the MAPP algorithm on a problem in SLIDABLE with $n$ units on a map of $m$ traversable tiles. We examine the worst case scenario in each case, and also discuss a best-case scenario at the end.

We introduce an additional parameter, $\lambda$, to measure the maximal length of alternate paths $\Omega$. In the worst case, $\lambda$ grows linearly with $m$. However, in many practical situations, $\lambda$ is a small constant, since the ends of an $\Omega$ path are so close to each other. Our analysis discusses both scenarios.

**Theorem 10.** *Algorithm 1 has a worst-case running time of $O(\max(n^2 m, m^2 \log m))$ when $\lambda$ is a constant, and $O(n^2 m^2)$ when $\lambda$ grows linearly with $m$.*

*Proof.* As outlined in Section 5.2, each *single-agent* A\* search with a consistent heuristic [2] expands $O(m)$ nodes. Hence, assuming that the open list is implemented as a priority queue, each A\* search takes $O(m \log m)$ time. Note that, on graphs where all edges have the same cost, the $\log m$ factor could in principle be eliminated using a breadth-first search to find an optimal path. Grid maps with only cardinal moves fit into this category. However, for simplicity, here we assume that the $\log m$ factor is present.

Hence, in the worst case, the searches for $\pi$-paths take $O(nm \log m)$ time for all $n$ units. The A\* searches for all $\Omega$'s take $O(m^2 \log m)$ time.

In a single progression step, outlined in Algorithm 2, suppose blank travel is required by all $n$ units, for every move along the way except the first and last moves. Since the length of $\pi$ paths is bounded by $m$ and the length of alternate paths $\Omega$ is bounded by $\lambda$, the total number of moves in a progression step is $O(nm\lambda)$, and so is the running time of Algorithm 2.

Clearly, the complexity of a repositioning step cannot exceed the complexity of the previous progression step. Hence the complexity of each iteration in Algorithm 1 (lines 5–7) is $O(nm\lambda)$. The number of iterations is at most $n$, since the size of $A$ reduces by at least one in each iteration. So MAPP takes $O(\max(nm \log m, m^2 \log m, n^2 m\lambda))$ time to run, which is $O(\max(n^2 m, m^2 \log m))$ when $\lambda$ is constant and $O(n^2 m^2)$ when $\lambda$ grows linearly with $m$. □

**Theorem 11.** *The maximum memory required to execute MAPP is $O(nm)$ when $\lambda$ is a constant, or $O(nm^2)$ when $\lambda$ grows linearly with $m$.*

---

2. It is well known that the Manhattan heuristic, which we used in our implementation, is consistent. The proof is easy, being a direct result of 1) the definition of consistency and 2) the way the Manhattan distance is computed (by pretending that there are no obstacles on the map).

*Proof.* Caching the possible $\Omega$ paths for the entire problem as described in Section 5.2 takes $O(m\lambda)$ memory. The A* searches for the $\pi$ paths are performed one at a time. After each search, $\pi$ is stored in a cache, and the memory used for the open and closed lists is released. The A* working memory takes only $O(m)$ space, and storing the $\pi$ paths takes $O(nm)$ space. Overall, path computation across all units requires $O(nm + m\lambda)$ space.

Then, in lines 5–7 of Algorithm 1, memory is required to store a stack of moves performed in one progression step, to be used during repositioning. As shown in the proof of Theorem 10, the number of moves in a progression step is within $O(nm\lambda)$. So, the overall maximum memory required to execute the program is $O(nm\lambda)$, which is $O(nm)$ when $\lambda$ is a constant and $O(nm^2)$ when $\lambda$ grows linearly with $m$. □

**Theorem 12.** *The total distance travelled by all units is at most $O(n^2m)$ when $\lambda$ is a constant, or $O(n^2m^2)$ when $\lambda$ grows linearly with $m$.*

*Proof.* As shown previously, the number of moves in a progression step is within $O(nm\lambda)$. The number of moves in a repositioning step is strictly smaller than the number of moves in the previous progression step. There are at most $n$ progression steps (followed by repositioning steps). Hence, the total travelled distance is within $O(n^2m\lambda)$. □

**Corollary 13.** *Storing the global solution takes $O(n^2m)$ memory when $\lambda$ is a constant, or $O(n^2m^2)$ when $\lambda$ grows linearly with $m$.*

We discuss now a best case scenario. MAPP computes optimal solutions in the number of moves when the paths $\pi$ are optimal and all units reach their targets without any blank traveling (i.e., units travel only along the paths $\pi$). An obvious example is where all paths $\pi$ are disjoint. In such a case, solutions are *makespan* optimal too. As well as preserving the optimality in the best case, the search effort in MAPP can also be smaller than that spent in a centralised A* search, being $n$ single-agent $O(m)$ searches, compared to searching in the combined state space of $n$ units, with up to $\frac{m!}{(m-n)!}$ states.

## 7. Extending the Completeness Range

To extend MAPP's completeness beyond the class SLIDABLE, we evaluated the impact of each of the three SLIDABLE conditions in a preliminary experiment. We ran Basic MAPP on the same data set that we also used in the main experiments (the data set and the main experiments are described in Section 9). In the preliminary experiment, we switched off one SLIDABLE condition at a time and counted how many units satisfy the remaining two conditions. A larger increase in the number of solvable units suggests that relaxing the definition of the condition at hand could provide a more significant increase in the completeness range.

This initial experimental evaluation indicates that Basic MAPP with all three SLIDABLE conditions solves 70.57% of units (basic case). If the alternate connectivity requirement is switched off, 87.06% units satisfy the remaining two conditions. Switching off target isolation makes 85.05% units satisfy the remaining two conditions. However, ignoring the blank availability condition has a very small impact, increasing the percentage only slightly, from 70.57% in the basic case to 70.73%. These results suggest that the alternate connectivity and the target isolation conditions are more restrictive than the blank availability condition. Thus, we focus on relaxing these two conditions.

For target isolation, our extension allows a unit to plan its path through other targets, when doing so can still guarantee that a clearly identified set of units will reach their targets. This is the topic of Section 7.1. To extend alternate connectivity, we developed a technique that allows $\pi$ paths to be planned through regions with no alternate paths, such as tunnels. The blank travelling operation for a tunnel-crossing unit now uses the blank positions ahead of this unit, along the remaining of its pre-computed path, as we describe in detail in Section 7.2. An empirical analysis of each of these features is provided in Section 9.

## 7.1 Relaxing the Target Isolation Condition

When several targets are close to each other, the target isolation condition, forbidding $\pi$ and $\Omega$ paths to pass through targets, can make targets behave as a "virtual wall", disconnecting two areas of the map. As a result, Basic MAPP can report many units as non-SLIDABLE.

The extension we introduce allows a unit $u$ to plan its path through the target of another unit $v$, if subsequently $v$ is never assigned a higher priority than $u$. More specifically, a partial ordering $\prec$ is defined, such that $u \prec v$ iff the target of $v$ belongs to the $\pi$-path of $u$ or to any $\Omega$ path along $u$'s $\pi$-path. Written more formally, $u \prec v$ iff $t_v \in \Pi(u)$, where $\Pi(u) = (\pi(u) \cup \bigcup_{i=1}^{k_u-2} \Omega_i^u)$. Every time we mention $\prec$ we refer to its transitive closure. We show in this section that, if paths can be planned in such a way that the (possibly empty) relation $\prec$ creates no cycles of the type $u \prec u$, then an instance can be solved with a slight modification of Basic MAPP.

Units plan their $\pi$ and $\Omega$ paths through a foreign target only when they have no other choice. To achieve this strategy, in the A* searches we assign a very high cost to graph search edges adjacent to a foreign target. This has the desirable outcome of reducing the interactions caused by the target isolation relaxation. In particular, the way the original SLIDABLE units compute their paths is preserved, as no foreign targets will be crossed in such cases. In other words, instances in SLIDABLE are characterized by an empty $\prec$ relation.

**Definition 14.** *An instance belongs to class* TI-SLIDABLE *iff for every unit $u$ there exists a path $\pi(u)$ satisfying the alternate connectivity and the initial blank condition as in the definition of* SLIDABLE *(Definition 1). Furthermore, no cycles are allowed in the $\prec$ relation.*

Assume for a moment that a (possibly empty) $\prec$ relation without cycles is available. Aspects related to obtaining one are discussed later in this section.

**Definition 15.** *For solving* TI-SLIDABLE *instances, the extended algorithm,* TI MAPP*, has two small modifications from the original algorithm:*

1. *The total ordering $<$ inside each progression step stays consistent with $\prec$: $u \prec v \Rightarrow u < v$.*

2. *If $u \prec v$, then $v$ cannot be marked as solved (i.e. moved from $A$ to $S$) unless $u$ has already been marked as solved.*

With these extra conditions at hand, we ensure that even if a unit $x$ arrives at its target $t_x$ before other units clear $t_x$ on their $\pi$-paths, those units can get past $x$ by performing the normal blank travel. Following that, $x$ can undo its moves back to $t_x$ in the repositioning step, as in Basic MAPP. To prove that TI MAPP terminates, we first prove the following two lemmas hold for the highest priority unit, $\bar{u}$, in any progression step:

**Lemma 16.** *No other unit will visit the target of $\bar{u}$, $t_{\bar{u}}$, in the current progression step.*

*Proof.* Since $\bar{u}$ is the master unit, it follows that $\bar{u} < v$ for any other active unit $v$. According to point 1 of Definition 15, it follows that $\bar{u} \prec v$. Relation $\prec$ has no cycles, which means that $v \nprec \bar{u}$. Therefore, by applying the definition of $\prec$, it follows that $t_{\bar{u}} \notin \Pi(v)$. This completes the proof, as in MAPP all movements are performed along $\pi$ and $\Omega$ paths. $\qquad\square$

Since a repositioning step can only undo moves made in the previous progression step, units will only revisit locations visited during that progression step. So, the following is a direct result of Lemma 16:

**Corollary 17.** *No other unit will visit $t_{\bar{u}}$ in the repositioning step that follows.*

**Corollary 18.** *After $\bar{u}$ is solved, it cannot interfere with the rest of the problem.*

**Theorem 19.** TI MAPP *terminates.*

*Proof.* Showing that at least the highest priority unit $\bar{u}$ reaches its target in a given progression step is virtually identical to the proof for Lemma 7. Corollary 18 guarantees that, after solving $\bar{u}$, it does not interfere with the rest of the problem. Hence, the number of active units strictly decreases after each progression step, and the algorithm eventually terminates. $\qquad\square$

Let us get back to the question of how to provide a cycle-free $\prec$ relation. Testing whether *all* units can plan their paths in such a way that no cycle is introduced might end up being expensive. When a unit $u$ can't possibly avoid all other targets, it might have to choose between crossing the target of $v$ or crossing the target of $w$. One option might lead to a cycle whereas the other might avoid cycles. Therefore, a systematic search might be required to seek a cycle-free relation $\prec$.

Rather than searching systematically, our TI MAPP takes a cheaper, greedy approach. If there are cycles, we mark a number of units as not being TI-SLIDABLE. These are selected in such a way that all other units remain cycle-free (we call these TI-SLIDABLE units). TI-SLIDABLE units are guaranteed to be solved.

As a result of its greedy approach, TI MAPP is not complete on class TI-SLIDABLE. Still, it is complete on a superset of SLIDABLE and it is able to identify many units (often all) that will be provably solved.

Finally, we wrap up the discussion of this extension by concluding that the upper bounds for MAPP, given in Section 6, still apply to TI MAPP. The proof is identical because we can make the same worst-case assumptions as before: only the master unit gets solved in a progression step, and every move along each unit's $\pi$ path requires blank travel. Moreover, note the additional step after path pre-computation for topologically sorting the partial order, $\prec$, into a linear priority order $<$, can be done cheaply in time linear on the number of units (Tarjan, 1976).

## 7.2 Relaxing the Alternate Connectivity Condition

As we will show in Section 9, the previous extension of target isolation significantly improves MAPP's *success ratio* (i.e., percentage of solvable units). Yet, there was significant room for further improvement. In particular, maps with single-width tunnels still showed a bottleneck in terms of success ratio. Tunnels make the alternate connectivity condition – connecting the two ends of a consecutive triple locations without going through the middle – harder or even impossible to satisfy. When a single-width tunnel bridges two otherwise disjoint regions, as shown in Figure 5, the versions of MAPP presented so far fail to find a path between the two regions, because alternate connectivity is broken for triples inside the tunnel.

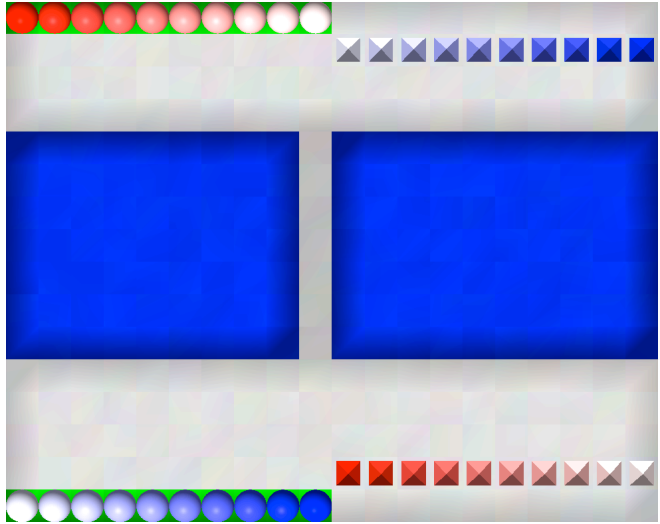

Figure 5: An example where all units have their targets on the other side, and the only way is to cross the single-width bridge. Units are drawn as circles, and their corresponding targets are squares of the same shade.

In this section we introduce the *buffer zone extension*, our solution for relaxing the alternate connectivity condition. It allows many paths, corresponding to many units, to cross the same single-width tunnel. The intuition is simple. Often, there are plenty of blank positions ahead of a unit, along the remaining locations of its precomputed $\pi$-path and the corresponding $\Omega$ paths along it. Our tunnel-crossing operation is essentially a generalisation of blank travelling, where a blank position is sought from the path ahead, instead of the alternate path of the current location triple.

**Definition 20.** *For each precomputed path $\pi(u)$ that crosses tunnels, we define the following:*

- *The* buffer zone*, $\beta(\pi(u))$, is the portion of $\pi(u)$ between the target and the end of the last tunnel (at the $j$-th move on $\pi(u)$), together with the corresponding alternate paths: $\beta(\pi(u)) = \bigcup\limits_{i \in \{j+2,\dots,k_u-1\}} \{l_i^u\} \cup \Omega_i^u.$*

- *A dynamic* counter*, $\kappa(\pi(u))$, keeps track of how many positions in the buffer zone are blank. The counter is initialized at the beginning with the appropriate value, and it is incremented and decremented as necessary later on.*

- *A* threshold *$\tau(\pi(u))$ is set to $\text{length}(\bar{t}) + 2$, where $\bar{t}$ is the longest tunnel to be crossed by the unit $u$. This threshold acts as a minimal value of $\kappa(\pi(u))$ that guarantees that $u$ can cross tunnels safely.*

When a unit $u$ attempts to cross a tunnel, it can push units of lower priorities to the closest blank locations in the buffer zone, until $u$ exits the tunnel. For such a tunnel-crossing operation to be possible, enough blanks have to be available in the buffer zone. Before we analyse the new extended algorithm in detail, we introduce the extended class Ac Slidable, whose definition includes those units meeting the new buffer zone extension.

**Definition 21.** *In relaxing the alternate connectivity condition, we allow $\pi(u)$ to go through one or more single-width tunnels iff there are enough blanks in $u$'s buffer zone, with at least $\tau(\pi(u))$ blank locations in $\beta(\pi(u))$ in the initial state, i.e., $\kappa(\pi(u)) \geq \tau(\pi(u))$. As before, alternate paths are still needed for all locations outside of tunnels.*

**Definition 22.** *A unit $u \in U$ belongs to the extended class, which we call Ac Slidable, iff it has a path $\pi(u)$ meeting the initial blank and the target isolation conditions as given in the definition of Slidable (Definition 1), and the relaxed alternate connectivity condition (Definition 21 above).*

Ac Mapp is modified from Basic Mapp in the following two ways to integrate our buffer zone technique for relaxing the alternate connectivity condition. Firstly, a repositioning step cannot finish if a counter $\kappa(\pi)$ has a value below the threshold $\tau(\pi)$. In other words, we need to ensure that enough blanks are available in a buffer zone before a progression step begins. The following is the new advancing condition, updated from Definition 2 by adding the extra, aforementioned condition.

**Definition 23.** *The* advancing condition *of an active, tunnel-crossing unit $u$ is satisfied iff its current position belongs to the path $\pi(u)$ and the next location on the path is blank (as given in Definition 2), and also $\kappa(\pi(u)) \geq \tau(\pi(u))$.*

Secondly, we need to preserve one of Basic Mapp's main features, where units of lower priority never block units with higher priority, ensuring that Mapp does not run into cycles or deadlocks. Hence, a unit $u$ with a lower priority than $v$ cannot *cause* moves that bring $\kappa(\pi(v))$ below the threshold (i.e., from $\tau(\pi(v))$ to $\tau(\pi(v)) - 1$). Recall that a move caused by $u$ is either a move of $u$ along its own $\pi(u)$ path (checked in lines 7-8 of Algorithm 4), or a move of a different unit $w$, which has been pushed around by $u$ as a side effect of blank travel (checked in lines 7-11 of Algorithm 3). Thus the buffer zone of $u$ acts as a generalised private zone, in which $u$ holds at least $\tau(\pi(u))$ locations that are not accessible to units with lower priorities.

Our extensions for Ac Mapp maintain the following properties of Basic Mapp.

**Lemma 24.** *As long as the master unit $\bar{u}$ is not solved, it is guaranteed to advance along $\pi(\bar{u})$ at each iteration of the outer ("while") loop in Algorithm 4. By the end of the current progression step, at least $\bar{u}$ has reached its target.*

*Proof.* Most of this result follows directly from the proof for Lemma 7. The parts that are new in Algorithm 4 compared to Algorithm 2 (the progression step in Basic Mapp) are the check in lines 7-8 and the modified blank travelling operation (lines 13-15). Since $\bar{u}$ has the highest priority of the current progression step, it can cause moves affecting the buffer zone of every other unit, but no other unit can move into the buffer zone of $\bar{u}$ when doing so would bring the number of blanks below the threshold, i.e. $\kappa(\pi(\bar{u})) < \tau(\pi(\bar{u}))$. Hence $\bar{u}$ is guaranteed to have enough blanks to cross each tunnel on $\pi(\bar{u})$. □

The proof for Lemma 25 below is very similar to Lemma 8 in Section 5.3.

---

**Algorithm 3** AC MAPP – `canBringBlank(` unit $u$, location $l_{i+1}^u$ )

---

1: **if** $u$ is outside of tunnels **then**
2:     look for a nearest blank $b$ along $\Omega_i^u$
3: **else if** $u$ is inside a tunnel **then**
4:     look for a nearest blank $b$ from $\beta(\pi(u))$
5: **if** no $b$ can be found **then**
6:     **return** false
7: **for** each location $l \in \{b, \ldots, l_{i+1}^u\}$ **do** {segment along $\Omega_i^u$ or $\beta(\pi(u))$ from above}
8:     **if** $\exists v < u : l \in \zeta(v)$ **then** {check if causing another unit to move into the private zone of a higher priority unit, $v$}
9:         **return** false
10:     **else if** $\exists v < u : l \in \beta(\pi(v))$ & $\kappa(\pi(v)) \leq \tau(\pi(v))$ **then** {check if causing another unit to move into the buffer zone of a higher priority unit, $v$}
11:         **return** false
12: **return** true

---

**Algorithm 4** AC MAPP – Progression step.

---

1: **while** changes occur **do**
2:     **for** each $u \in A$ in order **do**
3:         **if** pos$(u) \notin \pi(u)$ **then**
4:             do nothing
5:         **else if** $\exists v < u : l_{i+1}^u \in \zeta(v)$ **then**
6:             do nothing
7:         **else if** $\exists v < u : l_{i+1}^u \in \beta(\pi(v))$ & $\kappa(\pi(v)) \leq \tau(\pi(v))$ **then** {check when moving into the buffer zone of a higher priority unit}
8:             do nothing {wait until $v$ has more blanks in its buffer zone}
9:         **else if** $u$ has already visited $l_{i+1}^u$ in current progression step **then**
10:             do nothing
11:         **else if** $l_{i+1}^u$ is blank **then**
12:             move $u$ to $l_{i+1}^u$
13:         **else if** `canBringBlank(` $u$, $l_{i+1}^u$ ) **then** {Algorithm 3}
14:             bring blank to $l_{i+1}^u$
15:             move $u$ to $l_{i+1}^u$
16:         **else**
17:             do nothing

---

**Lemma 25.** *Algorithm 4 generates no cycles (i.e., no repetitions of the global state).*

**Theorem 26.** AC MAPP *terminates.*

*Proof.* It follows from Lemmas 24 and 25 that the number of active units strictly decreases over successive iterations of Algorithm 4. Hence, the algorithm AC MAPP eventually terminates.   □

Since we have shown that the algorithm AC MAPP is guaranteed to solve the class AC SLID-ABLE, the completeness result shown below follows directly.

**Corollary 27.** AC MAPP *is complete on the class* AC SLIDABLE.

The AC MAPP extension preserves the upper bounds on running time, memory usage, and solution length given in Section 6. Here, we introduce $\tau_{max}$ to denote the maximal length of tunnels that units have to cross. In our worst case analysis, all units initiate blank travelling for every move along the way, which now involves tunnels. So, depending on whether $\tau_{max}$ or $\lambda$, the maximal length of $\Omega$ paths, is longer, AC MAPP runs in $O(n^2 m \tau_{max})$ or $O(n^2 m \lambda)$ time. Since both parameters are often constant in practice, or grow at worst linear in $m$, the running time is $O(n^2 m)$ or $O(n^2 m^2)$, as before. The bounds for total travel distance and global solution follow directly. Lastly, there is virtually no additional memory required for storing the buffer zones, except for one counter and one threshold variable, per unit.

### 7.3 Combining Target Isolation and Alternate Connectivity Relaxations

We show that the two extensions to the SLIDABLE class can be combined.

**Definition 28.** *An instance belongs to the extended class,* TI+AC SLIDABLE, *iff for every unit $u$ there exists a path $\pi(u)$ meeting the initial blank condition as given in Definition 1, and the* relaxed alternate connectivity *condition from Definition 22. Furthermore, the (possibly empty) $\prec$ relation introduced as a result of target isolation relaxation has to be cycle-free, just as in Definition 14.*

We obtain an extended algorithm, TI+AC MAPP, by combining TI MAPP (Definition 15) and AC MAPP (Algorithms 3 and 4).

**Theorem 29.** TI+AC MAPP *terminates.*

*Proof.* As in the proof for Lemma 24, we can show that at least the highest priority unit $\bar{u}$ reaches its target in a progression step, as follows. From Definition 21, $\bar{u}$ is guaranteed to have enough blanks to clear through any single-width tunnels along its path. Definitions 14 and 15 guarantee that, outside of tunnels, $\bar{u}$ can always bring a blank when needed, as stated in Lemma 5. Furthermore, this progression step generates no cycles. This can be proved as in the cases of Lemmas 25 and 8.

We also know that the solved unit $\bar{u}$ does not interfere with the rest of the problem, from the results of Lemmas 16 and 18, and Corollary 17. Note that the tricky cases where units have their targets inside single-width tunnels are excluded from the extended class TI+AC SLIDABLE, because they have zero buffer capacity according to how $\beta$ was defined in Definition 20.

Since each iteration of the algorithm solves at least one unit, TI+AC MAPP terminates. □

## 8. Improving Solution Length

As mentioned before, to avoid replanning, units pushed off-track in blank travelling by other units undo some of their moves to get back on their $\pi$-paths in the immediate repositioning step. We have observed that, in practice, the reverse repositioning strategy (defined in Section 5.4) can introduce many unnecessary moves, which increase the solution length, increase the running time, and may hurt the visual quality of solutions.

Recall that, in a standard reverse repositioning step, new moves are added to the solution that is being built. These moves undo, in reverse order, moves of active units (i.e., those not solved yet) made in the previous progression step. The process continues until a well positioned state is reached, which means that all active units have their advancing condition satisfied (i.e., have every active unit on its $\pi$-path with a blank in front).

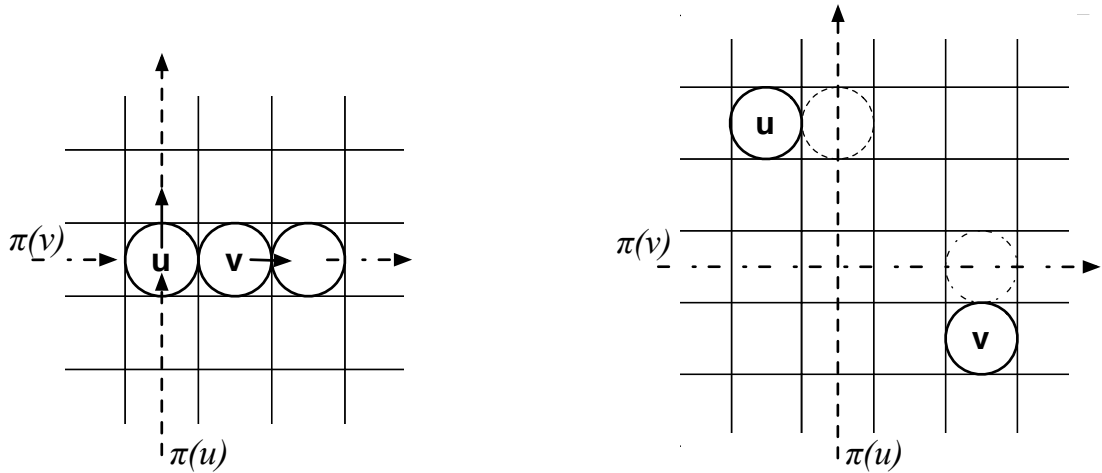

Figure 6: Two examples of global checking for a well positioned state.

After each undo move, the well-positioned check is performed globally. In other words, Basic MAPP checks the advancing condition for all active units, not only for the unit affected by the most recent undo move. This global checking guarantees to eventually reach a well-positioned state, as proved in Proposition 9, but, as mentioned earlier, it can often create many unnecessary moves.

We provide two simple examples in Figure 6, to illustrate one case when global checking is useful, and one case when global checking is too strong a condition, adding unnecessary moves. First, consider two units, $u$ and $v$, such that undoing one move off the global moves stack places $u$ back on its path, with a blank in front. Assume further that $u$'s current position is on the way of $v$'s future undo moves, as shown on the left of Figure 6. Therefore, even if $u$'s advancing condition is satisfied, $u$ needs additional undo moves, to make room for the undo moves of other units, such as $v$, in order to reach a globally well positioned state. In this case, global checking is useful. As a second example, imagine that $u$ and $v$'s moves in the most recent progression step are independent from each other, possibly even in two map areas far away from each other. A simple case is shown on the right of Figure 6. In the most recent progression, $v$'s last move (when $v$ was derailed) was followed by a sequence of $u$'s moves. Only the final move of $u$ pushed it off track, whereas the preceding moves were along $u$'s $\pi$-path, $\pi(u)$. Reverse repositioning would undo moves in the reverse *global* order, which means undoing all $u$'s moves before undoing $v$'s last move. However, only one undo move for $u$ and one undo move for $v$ are sufficient to restore both units to a well positioned state.

As we just illustrated, a global checking of the advancing condition could be too strong, whereas a local checking could be insufficient. The solution that we introduce in this section, which is called *repositioning with counting*, finds a middle ground between the two extremes, improves the number of moves and still maintains the guarantee of reaching a well-positioned state. Intuitively, the undo moves for a unit $u$ can stop as soon as (a) $u$'s advancing condition is satisfied, (b) its current position cannot possibly interfere with future undo moves of other units, (c) no other unit performing repositioning can possibly stop in the blank position in the front of $u$ on $u$'s $\pi$-path, and (d) $u$ doesn't stop in the "initial second location" of another active unit $v$. The initial second location of a unit $v$ is the position ahead of $v$ at the beginning of the most recent progression step. The fourth condition ensures that all units can have a blank in front at the end, as in the worst case they will revert back to the initial position at the beginning of the most recent progression step.[3]

**Definition 30.** *For each location $l$ on a $\pi$ or an $\Omega$ path, we keep a counter, $c(l)$, such that:*

- *At the beginning of each progression step, the counter $c(l)$ is reset to 0, if $l$ is empty, and to 1, if $l$ is occupied.*

- *Every time $l$ is visited during the progression step, $c(l)$ is incremented.*

- *Every time a unit leaves $l$ as a result of an undo move during the repositioning step, $c(l)$ is decremented.*

Following directly from the definition of $c(l)$ given above, we formulate the following two results for $c(l)$ at repositioning time:

**Lemma 31.** *If $c(l) = 0$, then no unit will pass through $l$ in the remaining part of the current repositioning step.*

**Lemma 32.** *For a given active unit $u$, at its current position $pos(u)$, if $c(pos(u)) = 1$, then all progression moves through the location $pos(u)$ have already been undone. In other words, no other unit in the remainder of this repositioning step will pass through $pos(u)$.*

We now introduce our new enhancement for MAPP, aimed at eliminating many useless undo moves in its repositioning steps.

**Definition 33.** *The enhanced algorithm, RC MAPP, uses the* repositioning with counting *strategy at line 7 of Algorithm 1. This means that each active unit $u$ stops undoing moves in the current repositioning step, as soon as it meets all of the following conditions:*

(a) *The advancing condition of $u$ is satisfied according to Definition 2, plus the extension in Definition 23.*

(b) *For $u$'s current location, $pos(u)$, $c(pos(u)) = 1$*

(c) *For the location in front of $u$, $l_{i+1}^u$, $c(l_{i+1}^u) = 0$*

(d) *The current location is not the initial second location of another active unit.*

**Theorem 34.** *All repositioning steps in RC MAPP end in a well-positioned state.*

---

3. Condition d) can be ignored without invalidating the algorithm's ability to make progress towards the goal state. Even if some units could possibly end up in a state without a blank in front, it is guaranteed that at least one unit (i.e., the one that finishes repositioning first) will have a blank in front. This further guarantees that at least one unit will be solved in the next progression step.

*Proof.* Recall that all the moves made in a progression step are kept as a totally ordered list. We can prove directly that repositioning with counting, by undoing a subset of those moves, reaches a well-positioned state. Since the counter $c(l)$ is incremented and decremented according to Definition 30, a unit $u$ satisfying all three conditions in Definition 33 has restored its own advancing condition. Furthermore, the combined results from Lemmas 31 and 32 guarantee that no other units will later get in $u$'s way, and that $u$ is 'out of the way' of the other units' repositioning moves. □

By Theorem 34, applying RC in the repositioning steps of our extended algorithm TI+AC MAPP has no negative impact on completeness.

## 9. Experimental Results

In this section we present our empirical evaluation of the MAPP algorithm. We first point out the impact of each newly added feature. Then we put our TI+AC+RC enhanced MAPP to the test in a comparison with existing state-of-the-art decoupled, incomplete methods. Specifically, our benchmarks are FAR (Wang & Botea, 2008), and an extended version of Silver's (2005) WHCA* algorithm by Sturtevant and Buro (2006), called WHCA*$(w, a)$, which applies abstraction to the expensive initial backward A* searches. As for MAPP, these algorithms have been tested on rather large problems, both in terms of map size and number of units. We are not aware of other programs that scale as well as FAR and WHCA*. The strengths of these two methods are the potential ability to find a solution quickly, but their weakness is that they cannot tell whether they would be able to solve a given instance.

We implemented MAPP from scratch and integrated it in the Hierarchical Open Graph[4] (HOG) framework. The source code for the extended WHCA* algorithm, WHCA*$(w, a)$ (Sturtevant & Buro, 2006), with the extra features of spatial abstraction and diagonal moves (but without a priority system for unit replanning), was obtained from Nathan Sturtevant. The FAR algorithm was our implementation as used in previous experiments (Wang & Botea, 2008).

Experiments were run on a data set of randomly generated instances used in previously published work (Wang & Botea, 2008). The input grid maps[5] are 10 of the largest from the game Baldur's Gate[6], which range from 13765 to 51586 traversable tiles in size, as listed in Table 1. These game maps are quite challenging, containing different configurations of obstacles forming different shapes of rooms, corridors, and narrow tunnels. We test each map with 100 to 2000 mobile units in increments of 100. A 10-minute timeout per instance is set. In the WHCA*$(w, a)$ experiments, we set the window size, $w$, to 8, and use the first level of abstraction ($a = 1$). This seems to be a good parameter setting in the work of Sturtevant and Buro (2006), and our experiments comparing with WHCA*(20,1) show WHCA*(8,1) to work better on this data set. Abstraction allows WHCA* to build the heuristic on a graph that is smaller than the actual graph where the movement takes place. In FAR, units make reservations for $k = 2$ steps ahead, which is the recommended setting. All experiments were run on a 2.8 GHz Intel Core 2 Duo iMac with 2GB of RAM.

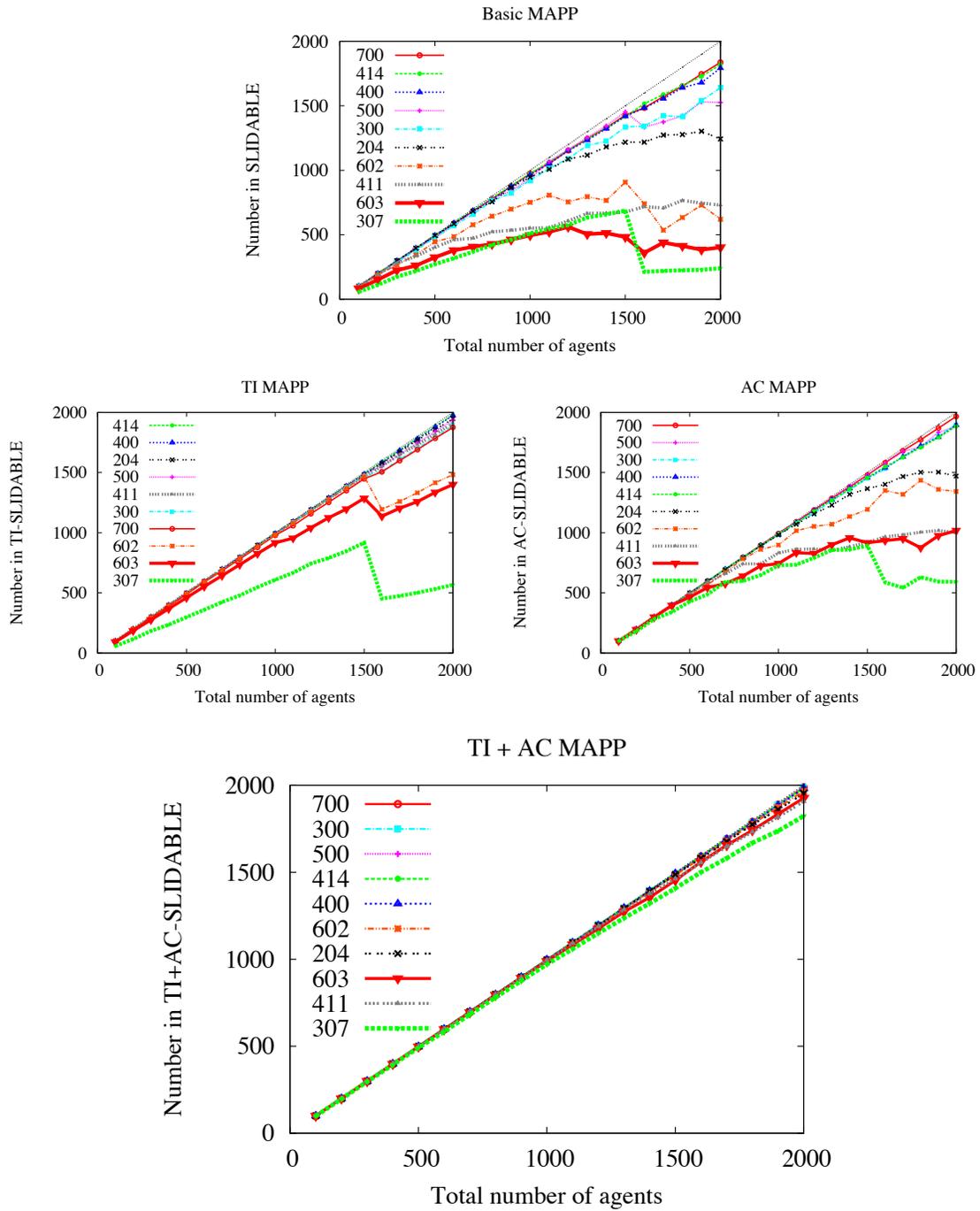

Figure 7: MAPP's widened completeness range after each relaxation: each graph line represents the number of units solved for problem instances on a map. Here, only provably solvable units are counted.

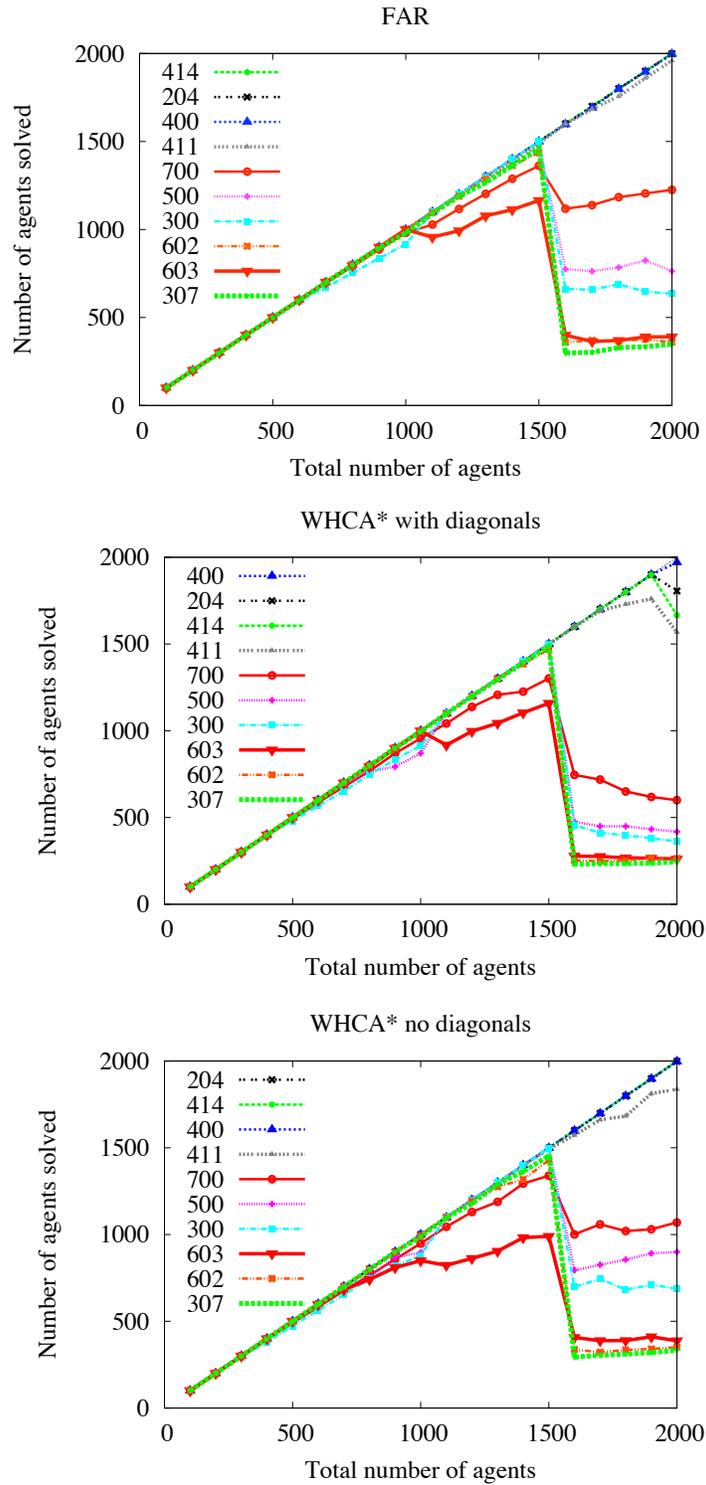

Figure 8: The success ratios (averaged over 10 trials) of FAR, and WHCA*(8,1), with and without diagonals, on the same set of problem instances. The timeout was set to 10 minutes per instance for all 3 incomplete algorithms.

| Map ID | Short ID | # nodes |
|--------|----------|---------|
| AR0700SR | 700 | 51586 |
| AR0500SR | 500 | 29160 |
| AR0300SR | 300 | 26950 |
| AR0400SR | 400 | 24945 |
| AR0602SR | 602 | 23314 |
| AR0414SR | 414 | 22841 |
| AR0204SR | 204 | 15899 |
| AR0307SR | 307 | 14901 |
| AR0411SR | 411 | 14098 |
| AR0603SR | 603 | 13765 |

Table 1: The 10 maps in descending order, in terms of number of nodes.

### 9.1 Scalability as Percentage of Solved Units

We compare FAR, WHCA*(8,1) and four versions of MAPP: Basic MAPP with the original SLIDABLE definitions; TI MAPP, the version with only the target isolation relaxation switched on; AC MAPP, based on relaxing the alternate connectivity condition; and TI+AC MAPP, relaxing both the target isolation condition, and the alternate connectivity condition. We measure the success ratio, which is defined as the percentage of solved units. Note that repositioning with counting (RC) does not have to be considered in this section, since it has no impact on the success ratio, being designed to improve solution length.

The MAPP versions used in this section attempt to solve only units that are provably solvable (i.e., units marked as SLIDABLE, TI SLIDABLE, AC SLIDABLE, TI+AC SLIDABLE respectively). The reason is that we want to evaluate how many units fall in each of these subclasses in practice. The next section will show data obtained with a version of MAPP that attempts to solve all units.

Figure 7 summarizes the success ratio data from each version of the MAPP algorithm on all maps. The closer a curve is to the top diagonal line (being the total number of units), the better the success ratio on that map. Basic MAPP exhibits a mixed behaviour, having a greater success ratio on six maps. On the four challenging maps (602, 411, 603, and 307), the success ratio gets often below 50% as the number of mobile units increases. These maps have a common feature of containing long narrow "corridors" and even single-width tunnels, connecting wider, more open regions of the map. Thus it is not surprising that, as mentioned in Section 7, the alternate path and target isolation conditions are identified as the greatest causes for failing to find a SLIDABLE path.

Relaxing the target isolation condition (TI MAPP) significantly improves the success ratio on all maps. Now a very good success ratio (93% or higher) is achieved for 7 maps across the entire range of the number of mobile units. The other 3 maps contain not only a high proportion of narrow corridors, but also single-width tunnels.

Relaxing alternate connectivity as well (TI+AC MAPP) yields an excellent success ratio for all unit numbers on all maps. For example, in the scenarios with 2000 units, which are the most challenging according to Figures 7 and 8, the smallest success ratio is 92% (map 307) and the largest one is 99.7%. In scenarios with fewer mobile units, TI+AC MAPP has even better success ratios.

---

4. http://webdocs.cs.ualberta.ca/~nathanst/hog.html

5. Our experimental maps can be viewed online, at: http://users.cecs.anu.edu.au/~cwang/gamemaps

6. http://www.bioware.com/games/baldurs_gate/

Next we compare the success ratio of TI+AC MAPP (bottom plot of Figure 7) to those of FAR (top plot of Figure 8) and WHCA*(8,1) (middle and bottom of Figure 8, with and without diagonals, respectively). Extended MAPP is the clear winner in terms of scalability. FAR and WHCA* suffer when the number of units is increased. These incomplete algorithms often time out even in scenarios with significantly fewer units than 2000. With 2000 units, FAR solves as few as 17.5% of units, while WHCA* solves as few as only 16.7% (no diagonal moves) and 12.3% (with diagonal moves) of the units. Over the entire data set, TI+AC MAPP solved 98.82% of all the units, FAR solved 81.87% of units, while 77.84% and 80.87% are solved by WHCA* with and without diagonal moves allowed, respectively.

## 9.2 Scalability when Attempting to Solve All Units

As in the previous section, we compare FAR, WHCA* and MAPP. The TI+AC MAPP version used here attempts to solve all units, not only the provably solvable ones (*attempt-all* feature). As mentioned earlier, this is achieved by marking all units as active at the beginning. Active units are partitioned into three categories: i) provably solvable units that did not reach their target; ii) other units that have not reached their target; and iii) units that have reached their target location, but they are still active because other units still have to cross through that location. The total ordering $<$ of active units must respect the conditions that units in category i) have a higher priority than units in category ii), which have a higher priority than units in category iii).

With the attempt-all feature turned on, TI+AC MAPP's percentage of solved units increases from 98.82% (Section 9.1) to 99.86%.

Next we focus on the number of solved instances. An instance is considered to be solved iff all units are solved. MAPP is successful in 84.5% of all instances. This is significantly better than FAR (70.6%), WHCA* with no diagonal moves (58.3%), and WHCA* with diagonals (71.25%).

The attempt-all feature has a massive impact on the percentage of fully solved instances, improving it from 34% to 84.5%. It might seem counter-intuitive that the attempt-all feature has a small impact on the percentage of solved units but a great impact on the percentage of solved instances. The explanation is the following. When MAPP fails in an instance, it does so because of a very small percentage of units that remain unsolved. Often, there can be as few as one or two unsolved units in a failed instance. Managing to solve the very few remaining units as well with the attempt-all feature will result in the whole instance changing its label from failed to solved, even though the change in the overall percentage of solved units will be small.

The remaining sections use the attempt-all feature as well. The reason is that it increases the number of solved instances and therefore we obtain a larger set of data to analyse.

## 9.3 Total Travel Distance

Factors that may impact the length of plans are the lengths of the initial $\pi$ paths, and the extra movements caused by blank travel and repositioning. In our experiments, the length of the precomputed $\pi$ paths has virtually no negative impact on the travel distance. Even when MAPP's $\pi$ paths have to satisfy additional constraints, such as avoiding other targets when possible, they are very similar in length with the normal unconstrained shortest paths, being only 1.4% longer on average.

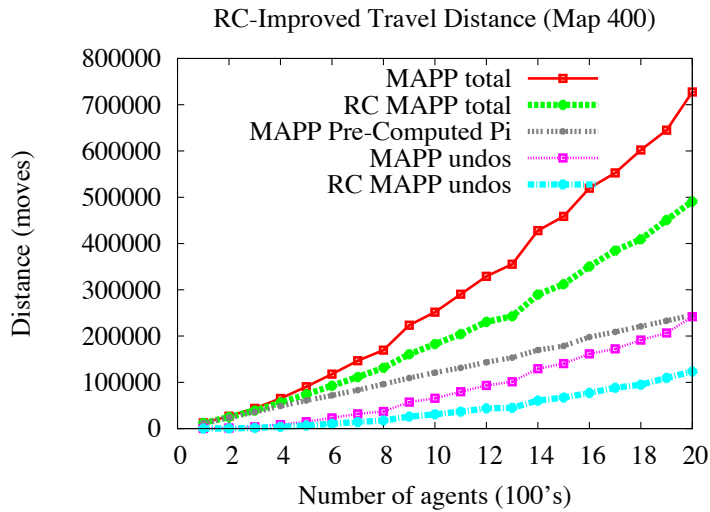

Figure 9: A typical case of improved distances of Rc Mapp over normal Mapp. Note the pre-computed $\pi$-paths are not affected by the Rc enhancement.

In this section, we first evaluate the improvement of repositioning with counting (Rc) over standard reverse repositioning. Then we compare the total distance travelled by Rc Mapp against Far and Whca*.

### 9.3.1 Reducing Undo Moves

We identified an excessive undoing of moves in repositioning as a bottleneck in Basic Mapp. Figure 9 shows the benefits of repositioning with counting (Rc), the enhancement described in Section 8. The figure compares the total travelled distance, as well as the number of undo moves, of Rc+Ti+Ac Mapp (shown as Rc Mapp for short) to Ti+Ac Mapp (Mapp for short) on an average case. As shown, repositioning with counting turns out to be quite effective, eliminating many *unnecessary* undo moves (that do not help to reach the globally restored state). Averaged over the entire data set, Rc Mapp has 59.7% *shorter* undo distance than Mapp with the standard reverse repositioning, which results in reducing the total travelled distance by 30.4% on average.

### 9.3.2 Comparing Total Distance with Far and Whca*(8,1)

Now we evaluate the solution length of attempt-all Rc+Ti+Ac Mapp as compared to Far and Whca*(8,1). We plot the total travel distance averaged over the subset of input instances that all algorithms considered here can fully solve.

Figures 10 and 11 show average results for all maps. For Mapp, we show the length of the precomputed $\pi$ paths, the number of the undo (repositioning) moves, and the total travelled distance. According to this performance criterion, the set of maps is roughly partitioned into three subsets.

In a good case, such as map 307, Mapp performs better than Whca*(8,1) without diagonals in terms of total travel distance, and is even comparable to Far. In an average case, Mapp's travel distance is roughly comparable to WHCA* without diagonals. Maps 603, 411, and 602 belong to this category. Finally, in a harder case, MAPP's total distance increases at a faster rate than the others, which is a direct result of an increasingly larger number of undo moves. These harder cases include maps 204, 414, 700, 400, 300, and 500. Upon inspection, these cases typically involve a high number of turns and corners. In MAPP's case, this results in a high degree of path overlapping, as units keep close to the edge when rounding a corner, to obtain shorter $\pi$-paths.

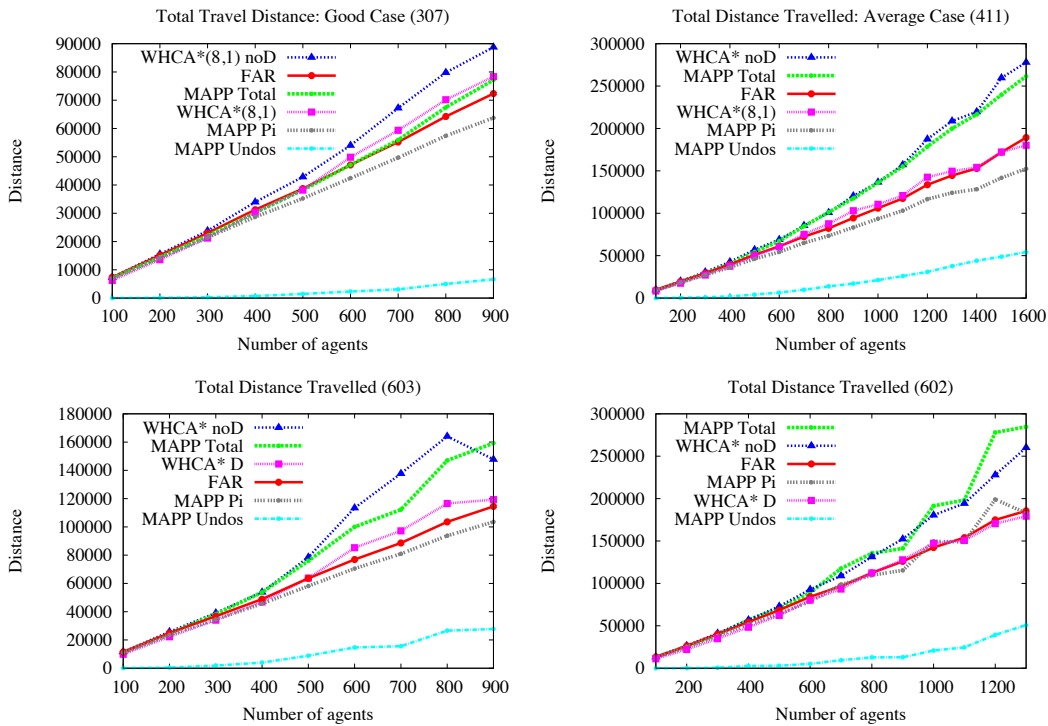

Figure 10: Distance travelled plotted averaged over instances fully solved by all algorithms.

To summarise the overall results, MAPP's travel distance ranges from 18.5% *shorter* than WHCA* without diagonal moves, to at most 132% longer, being 7% longer on average. Compared to the version of WHCA* with diagonal moves enabled, MAPP's total distance is 31% longer on average, varying from 5.8% *shorter* to at most 154% longer. Compared to FAR, MAPP's solutions range from 4.8% *shorter* to at most 153% longer, being 20% longer on average.

A closer look at the results reveals that, even with repositioning by counting in use, MAPP can still make unnecessary undo moves. Each useless undo move counts double in the final solution length, since the undo has to be matched with a new forward move in the next progression step. Improving the solution length further is a promising direction for future work.

## 9.4 Running Time Analysis

As in the case of travel distance analysis, for a meaningful runtime comparison, we also restrict the analysis to the subset of instances completed by all algorithms (FAR, both WHCA* versions, and T1+AC+RC MAPP with the attempt-all feature turned on). We show both overall summary data, in Tables 2 and 3, and charts for all 10 maps, in Figures 12 and 13.

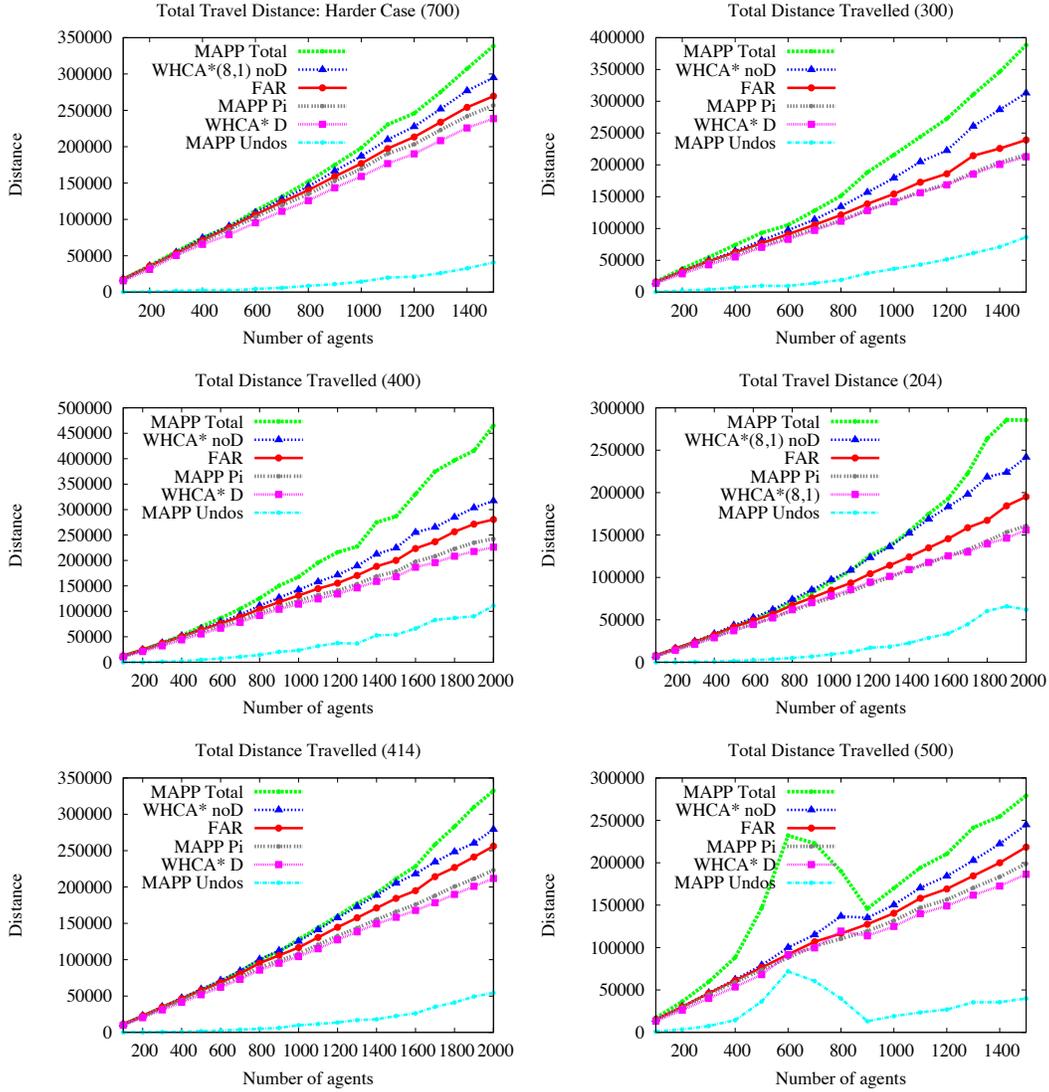

Figure 11: Distance travelled continued: the remaining six maps.

Our implementation of MAPP builds from scratch all the required paths, including the $\Omega$-paths. However, most $\Omega$-paths can be re-used between instances on the same map. Only $\Omega$-paths that contain a target in the current instance might have to be recomputed. This is a small percentage of all $\Omega$-paths, since the number of targets is typically much smaller than the map size. Such evidence strongly supports taking the $\Omega$-path computations offline into a map pre-processing step to improve MAPP's running time. Hence, we distinguish between the case when MAPP performs all computations from scratch, and the case when the alternate paths (i.e., $\Omega$ paths) are already available (e.g., from previous instances on the map at hand, or as a result of preprocessing). Note that with Far and Whca*, most computation depends on every unit's start and target locations, and therefore cannot easily be taken into a map pre-processing step (since storing entire search trees take up too much memory to be practical).

| Time ratio: | vs Far | vs Whca* | vs Whca*+d |
|---|---|---|---|
| Average | 10.14 | 0.96 | 0.93 |
| Min | 2.90 | 0.08 | 0.11 |
| Max | 60.46 | 4.57 | 4.92 |

Table 2: Mapp's runtime divided by the runtime of Far, Whca*, and Whca*+d. In this table, we assume that Mapp performs all computations, including alternate-path search, from scratch.

| Time ratio: | vs Far | vs Whca* | vs Whca*+d |
|---|---|---|---|
| Average | 2.18 | 0.21 | 0.19 |
| Min | 0.56 | 0.01 | 0.01 |
| Max | 7.00 | 0.99 | 0.70 |

Table 3: Mapp's runtime divided by the runtime of Far, Whca*, and Whca*+d. In this table, the time to compute alternate paths is omitted, as they could be re-used for instances on the same map.

Table 2 shows that, when Mapp performs all computations from scratch, it is comparable in speed with Whca*, being actually slightly faster on average. However, this version of Mapp is about 10 times slower than FAR on average. When $\Omega$ paths are already available, Mapp's speed improves significantly, as $\Omega$-path computation is the most expensive part of Mapp. As can be seen in Table 3, Mapp's speed ratio vs Far reduces to 2.18. Mapp also becomes 4.8–5.2 times *faster* than Whca*(with and without diagonals) on average.

Figure 12 shows more detailed runtime data for 8 out of 10 maps. Even when it does all computation from scratch, Mapp is faster than Whca*+d (i.e., with diagonal moves enabled). It is also often faster, or at least comparable, with Whca* without diagonals. Mapp with offline preprocessing is reasonably close to Far, even though Far is consistently faster or at least comparable to Mapp. The remaining two maps, which represent the most difficult cases for Mapp, are presented in Figure 13. On map 700 especially, the largest in our data set, which is also significantly larger than the rest (almost 2–4 times larger), Mapp has significantly higher total time, as shown at the top right of Figure 13.

A break down of Mapp's total running time (shown at the bottom of Figure 13 for map 700) consistently shows that the search time dominates. Furthermore, in node expansions, the $\Omega$ node expansions are generally several times greater than $\pi$ node expansions, resulting in the majority of path computation time being spent searching for $\Omega$-paths.

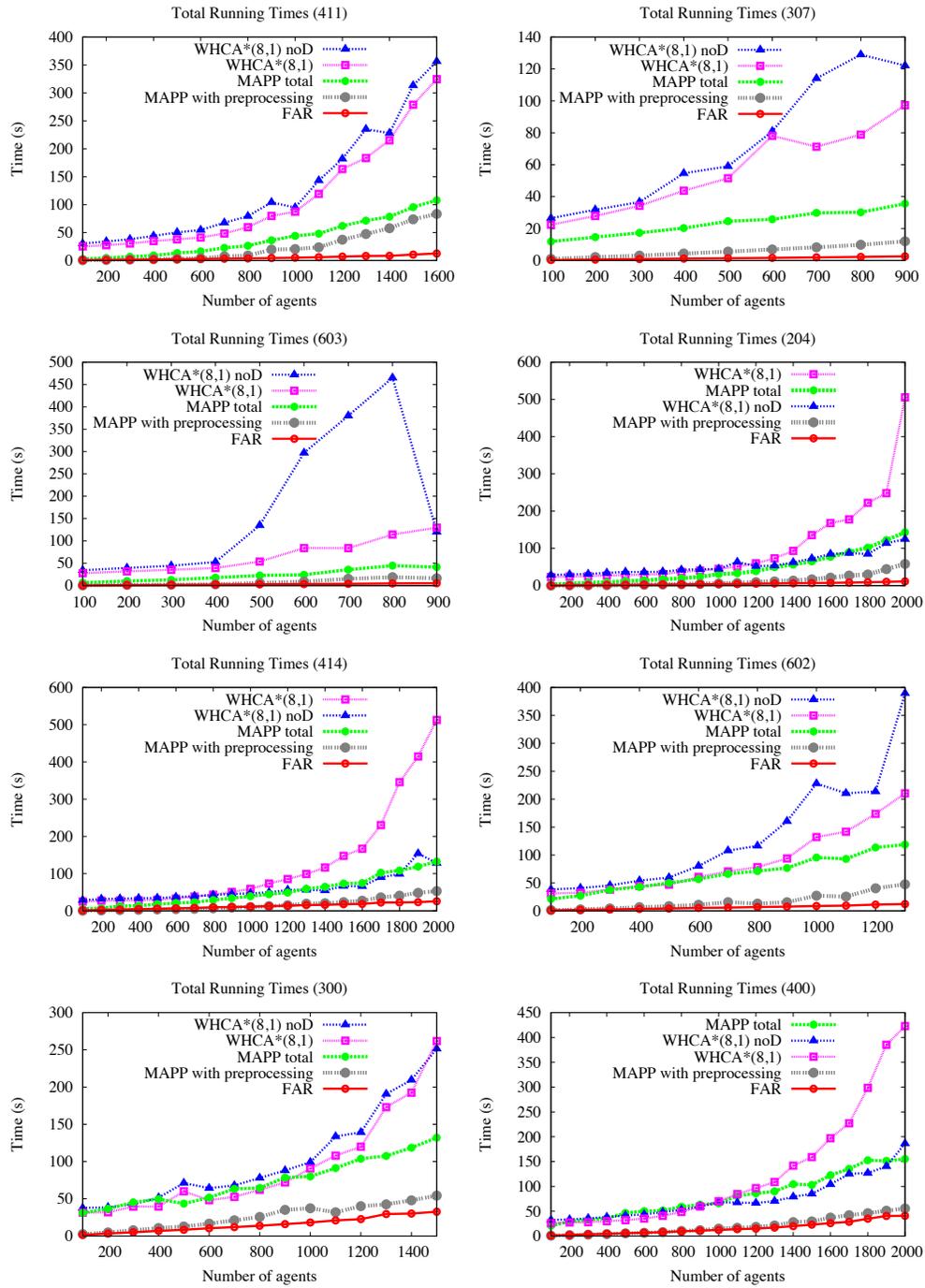

Figure 12: Runtime data averaged over fully completed instances by all algorithms. Map ID's are displayed in shorthand in brackets. "MAPP with preprocessing" stands for the version that computes no alternate paths.

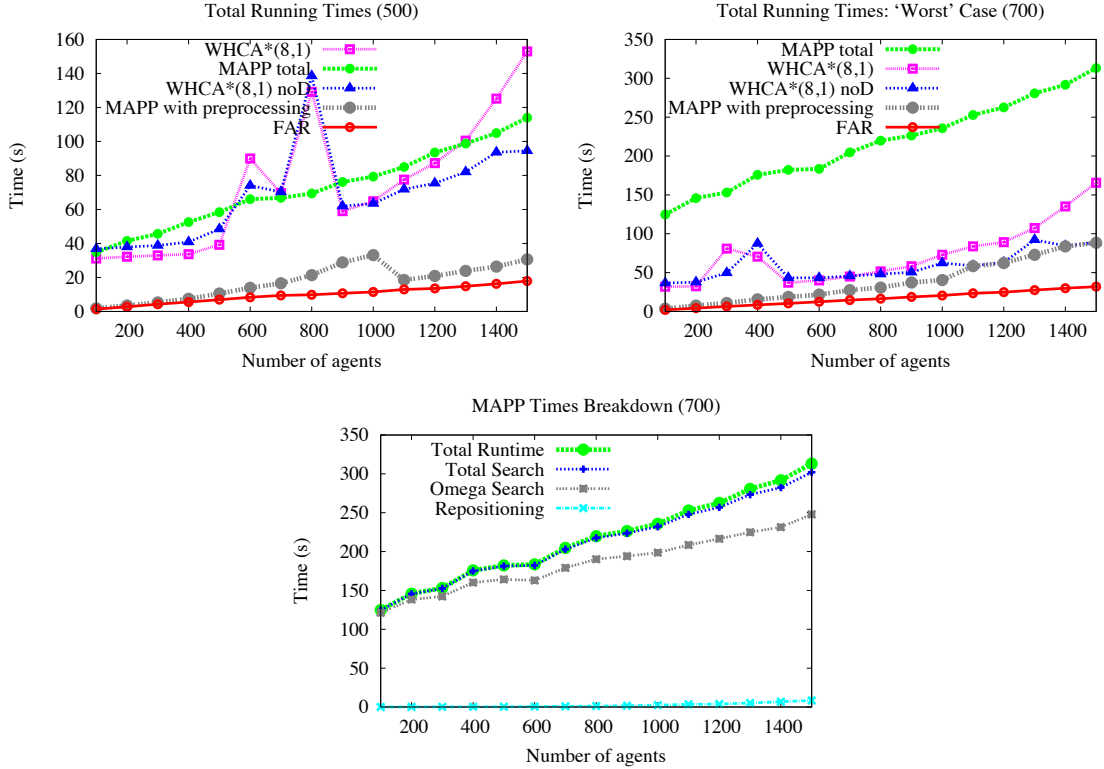

Figure 13: Top: hard cases for Ti+Ac+Rc Mapp's total runtime. Bottom: time breakdown, showing that the $\Omega$-path computation takes up the majority of Mapp's search time.

## 10. Conclusion

Traditional multi-agent path planning methods trade off between optimality, completeness, and scalability. While a centralised method typically preserves optimality and (theoretical) completeness, a decentralised method can achieve significantly greater scalability and efficiency. On the other hand, both approaches have shortcomings. The former faces an exponentially growing state space in the number of units. The latter gives up optimality and offers no guarantees with respect to completeness, running time and solution length. Our new approach, aimed at bridging these missing links, identifies classes of multi-agent path planning problems that can be solved in polynomial time. We also introduced an algorithm, Mapp, to solve problems in these classes, with low polynomial upper bounds for time, space and solution length.

We performed a detailed empirical evaluation of Mapp. The extended Mapp's completeness range reaches 92%–99.7%, even on the most challenging scenarios with 2000 mobile units. The completeness range is even better in scenarios with fewer units. On our data set, Mapp has a significantly better percentage of solved units (98.82% provably solvable, and 99.86% in the attempt-all mode) than Far (81.87%) and Whca* (77.84% and 80.87%, with and without diagonal moves). The attempt-all version of Mapp solves 13–26% more instances than the benchmark algorithms.

On instances solved by all algorithms, MAPP is significantly faster than both variants of WHCA*, and slower than the very fast FAR algorithm by a factor of 2.18 on average, when all alternate paths needed in an instance are readily available. When performing all computations from scratch, MAPP's speed is comparable to WHCA*. MAPP's solutions reported here are on average 20% longer than FAR's solutions and 7–31% longer than WHCA*'s solutions. However, unlike algorithms such as FAR and WHCA*, MAPP does offer partial completeness guarantees and low-polynomial bounds for runtime, memory and solution length. Thus, MAPP combines strengths from two traditional approaches, providing formal completeness and upper-bound guarantees, as well as being scalable and efficient in practice.

Our findings presented here open up avenues for future research into large-scale multi-agent pathfinding. In the long term, MAPP can be part of an algorithm portfolio, since we can cheaply detect when it is guaranteed to solve an instance. Thus it is worthwhile to investigate other tractable classes, such as subclasses where FAR is complete. MAPP can further be improved to run faster, compute better solutions, and cover more instances. Solution quality can be measured not only as total travel distance, but also in terms of makespan (i.e., total duration when actions can be run in parallel) and total number of actions (including move and wait actions). So far, we have worked on relaxing two of the original SLIDABLE conditions: target isolation and alternate connectivity. Future work could address the initial blank condition. Moreover, some of the initially non-SLIDABLE units in the problem could become SLIDABLE later on, as other SLIDABLE units are getting solved. Extending MAPP to instances where units are heterogeneous in size and speed is another promising direction.

## Acknowledgments

NICTA is funded by the Australian Government's Department of Communications, Information Technology, and the Arts and the Australian Research Council through Backing Australia's Ability and the ICT Research Centre of Excellence programs.

Many thanks to Nathan Sturtevant for providing the HOG framework, and his help with our understanding of the program. Thanks also to Philip Kilby, Jussi Rintanen, and Nick Hay for their many helpful comments. We thank the anonymous reviewers for their valuable feedback.